\PassOptionsToPackage{numbers}{natbib}
\documentclass[Afour,sageh,times]{sagej}

\usepackage[table,xcdraw]{xcolor}
\definecolor{colorfirst}{rgb}{0.866,0.945,0.831}   
\definecolor{colorsecond}{rgb}{1,0.98,0.83}         
\definecolor{colorthird}{rgb}{0.76,0.87,0.92}       
\definecolor{colorfourth}{rgb}{0.90,0.85,0.95}      

\newcommand{\cellfirst}{\cellcolor{colorfirst}}
\newcommand{\cellsecond}{\cellcolor{colorsecond}}
\newcommand{\cellthird}{\cellcolor{colorthird}}
\newcommand{\cellfourth}{\cellcolor{colorfourth}}

\usepackage[colorlinks=true, citecolor=red, linkcolor=red, urlcolor=red]{hyperref}

\usepackage{moreverb,url}
\usepackage{amsmath}
\usepackage{graphicx}
\usepackage{float}
\usepackage{pifont}
\usepackage{amssymb} 
\usepackage{subcaption}
\usepackage{makecell}
\usepackage{bm}
\usepackage{url}  
\usepackage{algorithm}
\usepackage{algorithmic}
\usepackage{multirow} 
\usepackage{booktabs}
\usepackage{pifont}  
\usepackage{tikz}
\usetikzlibrary{arrows, positioning}

\newcommand\BibTeX{{\rmfamily B\kern-.05em \textsc{i\kern-.025em b}\kern-.08em
T\kern-.1667em\lower.7ex\hbox{E}\kern-.125emX}}

\def\volumeyear{2025}

\begin{document}

\newcommand{\wzf}[1]{{\color{black}#1}}

\runninghead{Wang et al.}

\title{RoboBPP: Benchmarking Robotic Online Bin Packing with Physics-based Simulation}



\author{Zhoufeng Wang\affilnum{1}, Hang Zhao\affilnum{2, 3}, Juzhan Xu\affilnum{4}, Shishun Zhang\affilnum{1}, Ruizhen Hu\affilnum{4},   \\ Chenyang Zhu\affilnum{1}, Zecui Zeng\affilnum{5}, Weiyan Zhu\affilnum{6}, Zeyu Xiong\affilnum{1}, Haibin Yu\affilnum{2}, Kai Xu\affilnum{1,2}\textsuperscript{*}}

\affiliation{\affilnum{1}National University of Defense Technology, China, 
\affilnum{2}Institute of AI for Industries, Chinese Academy of Sciences,
\affilnum{3}Wuhan University, China, 
\affilnum{4}Shenzhen University, China,
\affilnum{5}JD Explore Academy, China,
\affilnum{6}China Post Technology Company, China}

\corrauth{Kai Xu, kevin.kai.xu@gmail.com
}


\begin{abstract}
Physical feasibility in 3D bin packing is a key requirement in modern industrial logistics and robotic automation. With the growing adoption of industrial automation, online bin packing has gained increasing attention. However, inconsistencies in problem settings, test datasets, and evaluation metrics have hindered progress in the field, and there is a lack of a comprehensive benchmarking system. Direct testing on real hardware is costly, and building a realistic simulation environment is also challenging. To address these limitations, we introduce RoboBPP, a benchmarking system designed for robotic online bin packing. RoboBPP integrates a physics-based simulator to assess physical feasibility. In our simulation environment, we introduce a robotic arm and boxes at real-world scales to replicate real industrial packing workflows. By simulating conditions that arise in real industrial applications, we ensure that evaluated algorithms are practically deployable. In addition, prior studies often rely on synthetic datasets whose distributions differ from real-world industrial data. To address this issue, we collect three datasets from real industrial workflows, including assembly-line production, logistics packing, and furniture manufacturing. The benchmark comprises three carefully designed test settings and extends existing evaluation metrics with new metrics for structural stability and operational safety. We design a scoring system and derive a range of insights from the evaluation results. RoboBPP is fully open-source and is equipped with visualization tools and an online leaderboard, providing a reproducible and extensible foundation for future research and industrial applications (\url{https://robot-bin-packing-benchmark.github.io/}).

\end{abstract}

\keywords{Bin Packing Problem, Robot Packing, Simulation, Benchmark, Dataset}

\maketitle

\section{Introduction}
The three-dimensional bin packing problem (3D-BPP) is a classical optimization problem with substantial importance in industrial logistics and warehouse management \citep{jqr-47-4-581, martello2000three,ha2017online}. In many industrial scenarios, items arrive sequentially, creating an online decision-making setting where the full sequence is unknown in advance \citep{seiden2002online}. Early efforts relied on heuristic methods \citep{crainic2008extreme,karabulut2004hybrid} to place items efficiently. With the rapid growth of industrial automation and robotic systems \citep{shome2019towards,wang2019stable,yang2021packerbot}, recent data-driven, learning-based approaches \citep{zhao2021online} have further advanced online packing methods, demonstrating strong performance across diverse industrial settings.

\begin{figure*}[t]
    \centering
    \includegraphics[width=0.9\textwidth]{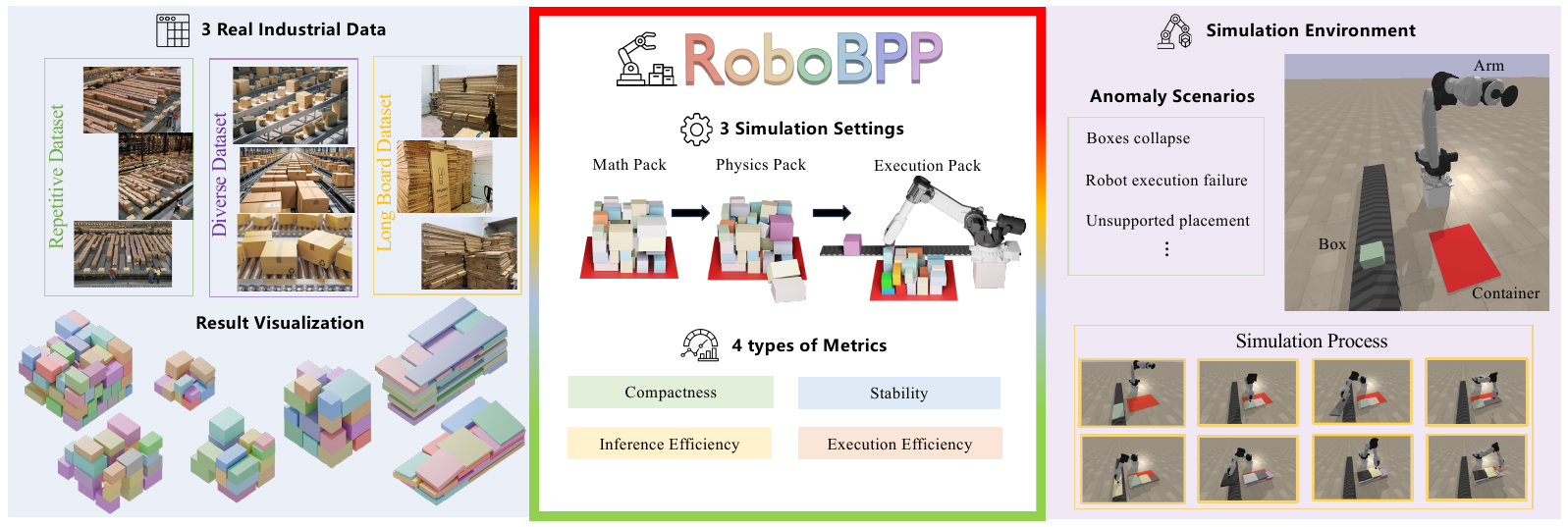}
    \caption{Overview of our benchmark for online 3D-BPP. We collect 3 real industrial datasets and build a physics-based simulation environment. We define 3 simulation settings and summarize 4 types of metrics.}
    \label{fig:intro}
\end{figure*}

\begin{table*}[t]
\centering
\caption{Comparison of representative datasets used in 3D bin packing research.}
\scalebox{0.75}{%
\begin{tabular}{lccccc}
\toprule
\textbf{Benchmark} & \textbf{Physical Feasibility} & \textbf{Data Type} & \textbf{Test Setting} & \textbf{Test Metric}\\
\midrule
\cite{martello2000three} & \ding{55} & Synthetic (simple) & Math & \ding{55} \\
\cite{elhedhli2019three} & \ding{55} & Synthetic (column-generation) & Math & \ding{55} \\
RS, CUT-1, CUT-2~\citep{zhao2021online} & \ding{55} & Synthetic (for DRL) & Math & \ding{55}  \\
BED-BPP~\citep{kagerer2023bed} & Rigid-body checks  & Real industrial (1 categories) & Math  & 2 dimensions\\
Ours & Simulation environment & Real industrial (3 categories) & Math, physics and execution & 4 dimensions \\
\bottomrule
\end{tabular}%
}
\label{tab:dataset_comparison}
\end{table*}

Despite these advances, the current research landscape remains fragmented. Existing works differ in problem settings, test data, and evaluation metrics, with many state-of-the-art methods still closed-source. Such inconsistencies hinder progress in the community, as no comprehensive benchmarking system exists to reliably assess algorithmic performance and real-world applicability. Direct evaluation on real hardware is often impractical, making a simulation environment essential for assessing the physical feasibility of algorithms. Unfortunately, most existing studies still treat 3D-BPP as a purely mathematical optimization problem, emphasizing compactness metrics such as Occupancy while overlooking physical constraints. Without modeling forces such as gravity, friction, and collisions, these methods may fail to perform reliably in real-world environments. Moreover, many prior studies rely on synthetically generated data whose distributions diverge from those of real industrial workflows, raising concerns about the practical validity of their findings. Recent efforts, such as BED-BPP~\citep{kagerer2023bed}, have begun to incorporate real order data and rigid-body checks, but they still assume idealized placements and do not continuously verify stability throughout execution.

Addressing this critical gap, we introduce RoboBPP (see Figure~\ref{fig:intro}), a comprehensive benchmarking system for robotic online bin packing. To assess physical feasibility, we develop a physics-based simulator that closely mirrors real industrial scenarios. The simulation environment incorporates a robotic arm and boxes at real-world scales, enabling a realistic reproduction of industrial packing workflows. The simulator continuously monitors stability during sequential placement and models realistic contingencies such as early collapses caused by unstable stacking and infeasible or unsafe robotic trajectories. In addition, robotic execution constraints are integrated to verify kinematic feasibility and operational safety, ensuring that the benchmark faithfully reflects real industrial operations.

Beyond physical realism, a comprehensive benchmark also requires scientific rigor in its datasets, test settings, and evaluation metrics. We provide three large-scale datasets that are both realistic and diverse (Figure~\ref{fig:data} visualizes the datasets). Our simulation framework includes three carefully designed test settings—from purely geometric evaluation to physics-constrained simulation and full robotic execution—enabling systematic assessment under increasingly realistic conditions (Figure~\ref{fig:test setting} illustrates these settings). The benchmark incorporates multi-dimensional metrics, including compactness, stability, inference efficiency, and execution efficiency. In addition, we design a scoring system that normalizes all metrics to a unified scale and computes a weighted overall score, enabling fair and holistic comparison across algorithms.

Alongside establishing a comprehensive benchmarking system, we reproduce a wide range of existing methods. Our evaluation yields practical insights for industrial deployment, such as identifying which classes of algorithms are better suited to specific application scenarios. For example, in highly repetitive assembly-line production environments, RL algorithms that explicitly model spatial and geometric relationships—such as PCT~\citep{zhao2021learning} and TAP-Net++~\citep{xu2023neural}—perform particularly well. In logistics scenarios with substantial variation in item sizes, RL methods equipped with Transformer-based packing policies—such as PCT and AR2L~\citep{pan2023adjustable}—are more effective, as they can learn diverse size-distribution patterns. In scenarios dominated by elongated furniture items, both RL algorithms and geometry-driven heuristics—such as TAP-Net++ and DBL~\citep{karabulut2004hybrid}—are especially suitable. Furthermore, we conduct a cross-dataset and cross-setting performance comparison, a single-metric analysis, and an investigation of stability-related metrics, providing deeper insights into algorithmic generalization and robustness across industrial environments.

In summary, the core contributions of this work include: 1) A highly realistic physics-based simulation environment for evaluating physical feasibility; 2) Three large-scale, diverse datasets collected from real industrial workflows for benchmarking online bin packing; 3) Scientifically designed test settings, ranging from purely mathematical evaluation to physics-constrained simulation and full robotic execution; and 4) Multi-dimensional evaluation metrics, combined with a scoring system that computes a weighted overall score, enabling insightful analysis of different approaches across various scenarios.

\section{Related Research}
We discuss recent developments in benchmarking systems and introduce the algorithms evaluated in this work.

\subsection{Development of Benchmark System}
We evaluate benchmark systems along four dimensions: physical feasibility, the realism and diversity of the data, test settings, and evaluation metrics (as shown in Table~\ref{tab:dataset_comparison}).

Early 3D-BPP research focused on algorithm design and relied on instance generators that produce items within predefined ranges. For example, \citep{martello2000three} created instances across nine categories with varying container and item sizes. While these increased diversity, the complexity remained uncontrolled, and rules were simple, limiting relevance to modern research. \cite{elhedhli2019three} noted the lack of realistic benchmarks and proposed a column-generation-based instance generator to generate test instances. With the growing popularity of learning-based solvers, \cite{zhao2021online} constructed three datasets, RS, CUT-1, and CUT-2, representing different sequence generation strategies for training DRL models. However, these datasets had fixed container dimensions and limited item types. These early benchmark systems were mainly focused on datasets and did not provide environments for evaluating physical feasibility, diverse test settings, or multidimensional evaluation metrics.

\cite{kagerer2023bed} introduced BED-BPP from real order records, with rich item information and stability evaluation via rigid-body simulation. It brings research closer to real industrial conditions. However, it only assesses the final placement stability and assumes idealized placement, without robotic execution. Its metrics focus on geometric compactness and stacking stability, and it is only tested under a purely mathematical setting.

We provide three large-scale datasets that are both realistic and diverse. Our system features a physics-based simulation environment to assess physical feasibility, incorporating robotic placement and continuous stability checks for realistic evaluation. Furthermore, the simulation includes three carefully designed test settings and employs a multi-dimensional metric system that measures compactness, stability, inference efficiency, and execution efficiency, enabling comprehensive and industrially aligned performance assessment.

\subsection{Algorithmic Progress}

In real industrial scenarios, boxes arrive one by one and require immediate placement. Online settings have gained substantial attention driven by industrial demands, and both heuristic and learning-based online algorithms have been proposed. We divide these methods into four categories and describe them in the following sections.

\textbf{Rule-based Greedy Heuristics}, including DBL~\citep{karabulut2004hybrid}, LSAH~\citep{hu2017solving}, HM~\citep{wang2019stable}, and SDFPack~\citep{pan2023sdf}, represent single-rule greedy strategies that rely on explicit geometric or cost functions, greedily selecting the optimal feasible placement according to a fixed criterion (e.g., the lowest-left position or the minimum surface area). Their placement strategies are generally conservative, which in turn leads to favorable performance in real-world deployments. Among them, the deep-bottom-left (DBL) heuristic \citep{karabulut2004hybrid} remains highly influential due to its simplicity and robust performance, placing each item in the lowest and most-left feasible position. \cite{hu2017solving} proposed the Least Surface Area Heuristic (LSAH), a least-surface-area-based strategy that examines all empty maximal spaces and six orientations of the item to identify the placement minimizing the total occupied surface area. \cite{wang2019stable} developed the Heightmap-Minimization (HM) method, selecting placements that minimize the increase in occupied volume, thereby improving stacking compactness. More recently, \cite{pan2023sdf} developed SDFPack, employing signed distance field representations to characterize container occupancy and guide placements toward minimal geometric conflict and better stability.

\textbf{EMS-driven Heuristics}, such as OnlineBPH~\citep{ha2017online}, MACS~\citep{hu2020tap}, and BR~\citep{zhao2021online}, explicitly maintain and evaluate empty maximal spaces (EMS) lists to search for the optimal placement among all feasible spatial configurations. They are generally used to fulfill specific strategy roles within learning-based methods. \cite{ha2017online} introduced the online packing heuristic (OnlineBPH), which also follows the DBL order and evaluates all possible combinations of bins, items, EMS, and rotations across opened containers before selecting the optimal placement. \cite{hu2020tap} introduced the Maximize-Accessible-Convex-Space (MACS) heuristic, which selects the candidate location that maximizes the remaining usable space. \cite{zhao2021online} proposed the BR heuristic, which is inspired by the observation that humans tend to maintain a regular packed volume to preserve larger remaining spaces, and evaluates each packing action by the regularity of the resulting bin.

\textbf{Rule-guided RL Methods}, such as PackE~\citep{yang2021packerbot} and CDRL~\citep{zhao2022learning}, integrate heuristic priors or reward shaping to guide the learning process toward stable and efficient placements. These approaches suit high-risk or high-precision tasks but tend to incur additional computational overhead and exhibit limited generalization. Early methods include PackE~\citep{yang2021packerbot}, which combines multiple heuristics to generate candidate placements and applies negative rewards to discourage positions far from preferred locations, guiding the agent to select feasible and stable placements. CDRL~\citep{zhao2022learning} formulates packing as a sequential decision problem using an actor-critic DRL framework, performing fast online stability analysis, decoupling policy along length, width, and orientation, and favoring far-to-near placement to simplify collision-free execution. 

\textbf{Geometry-aware RL Methods}, including PCT~\citep{zhao2021learning}, TAP-Net++~\citep{xu2023neural}, and AR2L~\citep{pan2023adjustable}, focus on modeling spatial and geometric relationships among items or candidate placements to enhance decision precision and adaptability. They are capable of learning structural patterns in the current scene, exhibiting strong generalization and adaptability, and most of them are built upon Transformer architectures. The breakthrough came with PCT~\citep{zhao2021learning}, the first learning-based method for online 3D-BPP in a continuous solution space. Instead of directly regressing coordinates, it identifies feasible candidates from the continuous domain and uses DRL to select the optimal one, balancing precision and stability \citep{yuan2023towards}. TAP-Net++~\citep{xu2023neural} extends TAP-Net~\citep{hu2020tap} with Cross-Modal Transformer (CMT) to compute attention between multiple candidates and multiple items, enabling dynamic adaptation to box shape. It also introduces EMS decomposition for real-time feasibility checking. Most recently, AR2L~\citep{pan2023adjustable} has introduced adjustable RL, adaptively balancing average and worst-case performance by tuning constraint weights during training, improving robustness under uncertain item distributions and physical constraints.

\begin{figure*}[t]
\centering
\includegraphics[width=0.95\textwidth]{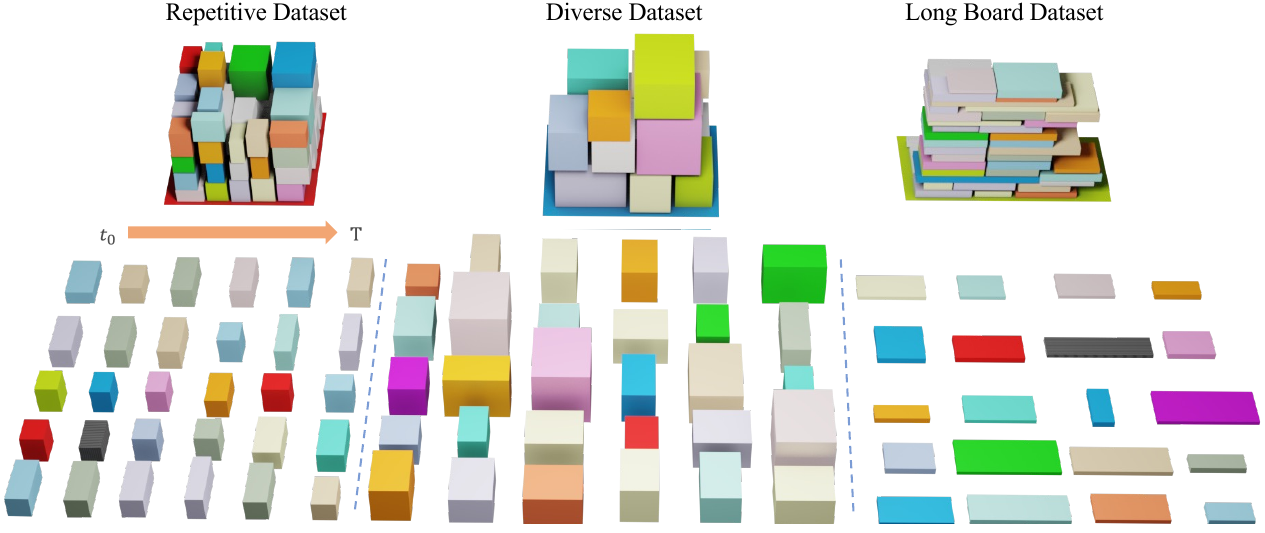}
\caption{Visualization of the three industrial datasets. Each block contains two parts: the upper row shows the packing results inside a container, while the lower row displays the corresponding item distribution and their tasks.}
\label{fig:data}
\end{figure*}

\section{RoboBPP Benchmark} 
This section introduces the components of our benchmark system, including the simulation environment, the dataset, the test setting, and the evaluation metrics. All resources are publicly available on our open-access website.

\subsection{Simulation Environment}

Since direct testing on real hardware is costly and operationally difficult, we build a simulation environment to evaluate physical feasibility. The core difficulty lies in constructing a sufficiently realistic simulator, which is essential for ensuring that evaluated algorithms function reliably in the real world. Inspired by actual industrial packing processes, we introduce a robotic arm with real-world scale and design a set of physically grounded parameters in a PyBullet-based environment, allowing both the boxes and the robotic arm to be reproduced in simulation. \wzf{To maintain numerical stability, the simulator parameters (e.g., timestep, friction coefficients, and contact dynamics) are set within physically reasonable ranges. Unrealistic parameter combinations can cause excessive collision responses and unstable object motions. Therefore, we adopted moderate and widely adopted settings and verified through repeated simulation runs that objects remained stable.} Our simulator can reproduce various real-world conditions: for example, it incorporates gravity and friction to simulate collapse caused by unstable stacking, uses the OMPL motion planning library~\citep{sucan2012open} to generate robot trajectories, and enables the robotic arm to perform box gripping and placement operations. Compared with the rigid-body stability checks and idealized placement assumptions in BED-BPP~\citep{kagerer2023bed}, RoboBPP offers advantages in robot execution and continuous stability monitoring, enabling it to more closely approximate real industrial operating conditions.

Specifically, we build our simulation environment in PyBullet, using an ABB IRB 6700-200/2.60 industrial robot~\citep{abbirb6700} equipped with a suction gripper. \wzf{The gripper uses a simplified attachment mechanism to keep picked objects bound to the suction cup during transport and placement. This approach approximates industrial suction behavior without explicitly simulating suction forces, which is consistent with our benchmark design. Following existing work~\citep{zhao2025deliberate}, we assume a suction cup with a maximum force of 260\,kg/m$^2$, under which object dropping does not occur under normal manipulation.} The robot performs grasping and placement under full physical constraints. A red container is mounted on a wooden table, while a conveyor belt continuously delivers online-generated boxes, as shown on the right side of Figure~\ref{fig:intro}. The environment is fully compatible with OpenAI Gym. The entire simulation environment is released as an open-source Python package, \href{https://pypi.org/project/packsim/}{packsim}, on PyPI. To run the simulation, users follow the instructions in our \href{https://robot-bin-packing-benchmark.github.io/documentation.html}{online documentation}. When using the simulation environment, test data are loaded in an online manner, and the selected packing algorithm determines the placement location for each incoming box. This placement is then mapped into the simulation environment, where the robotic arm performs the corresponding grasping and placement actions. The process continues until all boxes are successfully placed or a collapse occurs.

\subsection{Dataset}
\begin{figure}[t]
    \centering
    \includegraphics[width=0.45\textwidth]{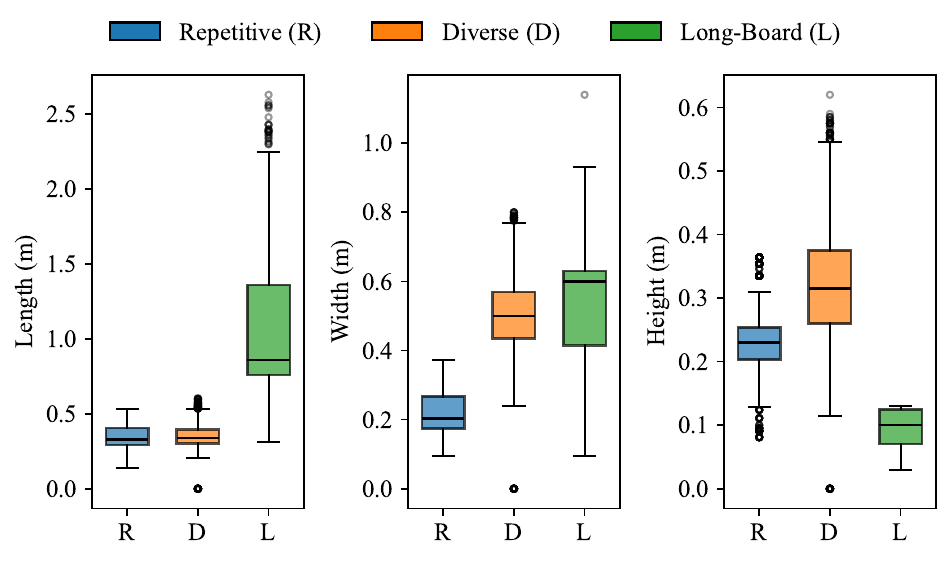}
    \caption{Box plots of item dimensions (length, width, height) for the three datasets.}
    \label{fig:boxplot}
\end{figure}

For a comprehensive benchmark, it is essential to cover diverse real-world scenarios. We analyzed common industrial workflows and identified three representative task scenarios (Figure~\ref{fig:data}). The first scenario involves assembly-line production, featuring highly repetitive boxes. We obtained order data from a household goods manufacturer, forming our Repetitive Dataset. The second scenario pertains to logistics packing, characterized by a high variation in box sizes. The high variability and volume of the data make statistical characterization difficult. We collected order data from an office supplies vendor and calculated the proportion of each box category, forming our Diverse Dataset. The third scenario involves packing items with unusual shapes, which makes placement particularly challenging, such as elongated panels. We obtained order data from a furniture manufacturer, forming our Long Board Dataset. Figure~\ref{fig:boxplot} summarizes the item size distributions of the three datasets in boxplots.

\begin{figure}[tbp]
    \centering
    \includegraphics[width=0.47\textwidth]{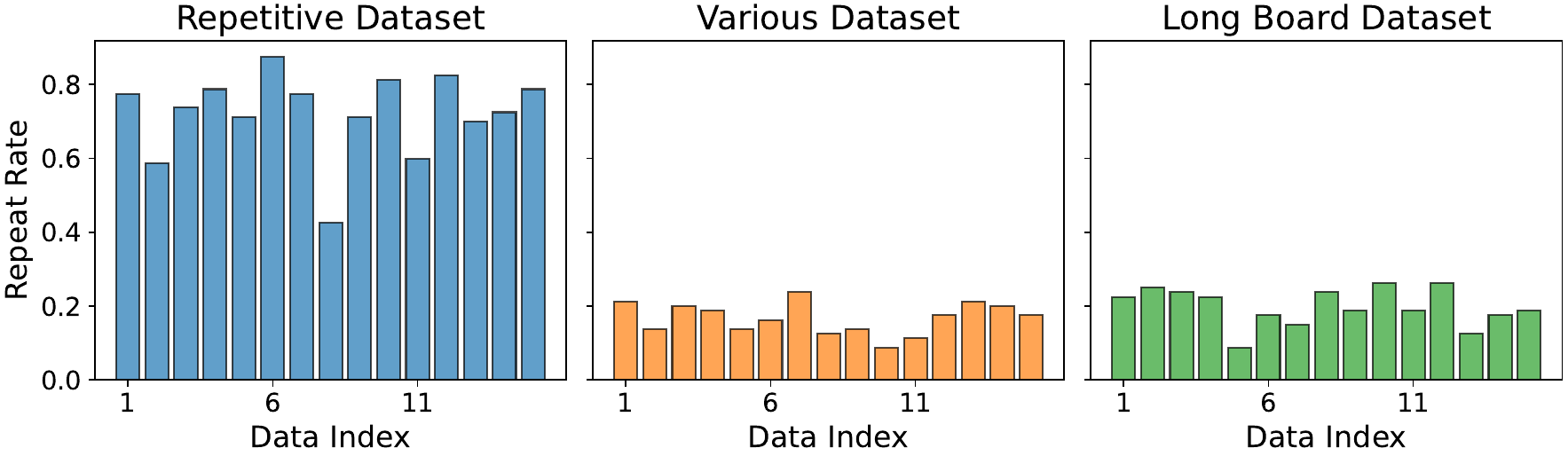}
    \caption{Repeat rates of 15 randomly selected groups for the three datasets, with each bar representing the proportion of duplicate boxes in that group.}
    \label{fig:repeat rate}
\end{figure}

\textbf{Repetitive Dataset} features consecutively repeated boxes observed in real logistics operations, where identical items are produced and packed in continuous sequences, similar to assembly-line manufacturing. It is designed to evaluate whether an algorithm can recognize repetitive patterns and maintain stable placement over time. Such repetition is common in industrial production and supply workflows. Figure~\ref{fig:repeat rate} shows the repeat rates of 15 randomly selected groups from our datasets, where the repeat rate is defined as the number of repeated boxes divided by the total number of boxes in each group. \wzf{The dataset contains 16,767 time-ordered records, where each record includes item dimensions, mass, and timestamp information.} It characterizes a repetitive packing scenario that emphasizes minimizing cumulative errors and ensuring stable stacking during continuous online packing.

\textbf{Diverse Dataset} features a large variety of boxes with differences in size, aiming to evaluate an algorithm’s adaptability to highly diverse item distributions. Such heterogeneity is common in logistics order data, from which the proportion of different item categories can be derived to reflect realistic demand distributions. \wzf{It comprises 6,849 office supply items with dimensions, mass, and proportion metrics.} This dataset represents a diverse task that tests an algorithm’s ability to manage diverse items while optimizing both space utilization and stacking stability.

\textbf{Long Board Dataset} features boxes with one dimension longer than the others, aiming to evaluate an algorithm’s ability to handle geometrically challenging shapes. Such items are common in manufacturing and furniture logistics, where their geometry complicates spatial arrangement and stability. As shown in Figure~\ref{fig:boxplot}, the Long Board Dataset exhibits a notably wider range in length compared to the other datasets, reflecting its elongated, panel-like nature. \wzf{Each record includes item dimensions and mass information.} This dataset represents a long-board–packing task requiring precise control of positioning and balance, thus providing a rigorous test of algorithm robustness under physically constrained conditions.

\begin{figure*}[t]
\centering
\includegraphics[width=0.95\textwidth]{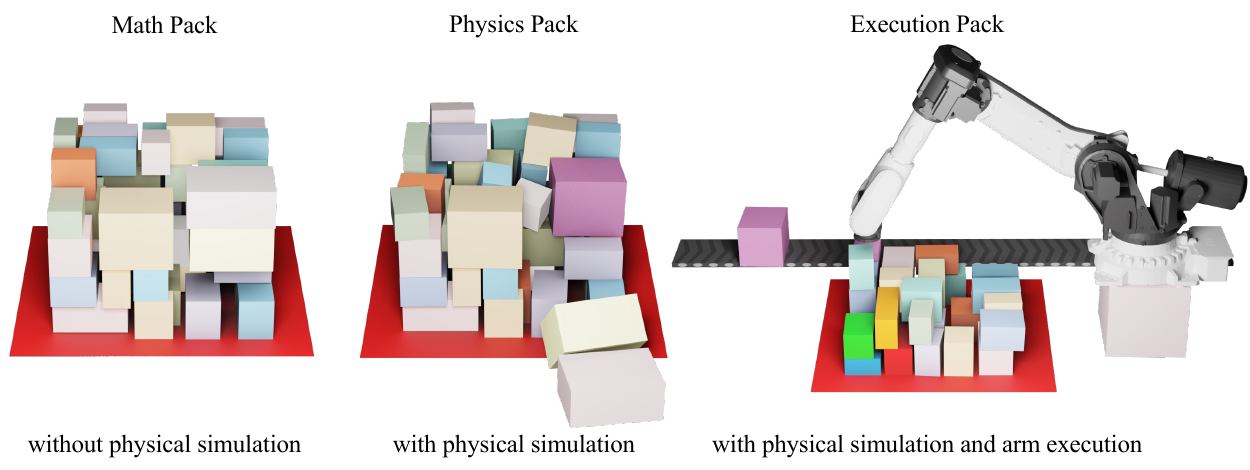}
\caption{Three test settings in the simulation environment. \textbf{Math Pack} evaluates purely geometric placement. \textbf{Physics Pack} adds gravity and collision test under realistic physical constraints. \textbf{Execution Pack} integrates physics and robotic execution for end-to-end evaluation.}
\label{fig:test setting}
\end{figure*}

\textbf{Dataset Usage:} The collected real-world data are divided into training and testing subsets. The training sets are publicly available for algorithm development, while the testing sets remain private and are used exclusively for leaderboard evaluation. \wzf{The datasets are constructed primarily based on item dimensions and mass.} The collapse threshold specifies the maximum allowable positional deviation for a box; exceeding this value is considered a collapse. \wzf{This threshold was determined through repeated simulation experiments rather than chosen heuristically. Specifically, we tested a range of candidate threshold values in the simulator and manually inspected the corresponding placement results. During this process, we examined whether the observed motion after placement corresponded to normal local settling or to genuine instability, such as obvious box displacement, tilting, or chain-reaction disturbance of previously placed items. The final threshold was selected as the smallest value that consistently captures clear collapse cases while avoiding misclassification of minor acceptable settling motions. In this way, the chosen threshold reflects practically meaningful collapse events that are likely to be considered failures in real industrial packing scenarios.} The \emph{Repetitive Dataset} uses a container size of $(1.34 \text{m},1.25\text{m},1.00\text{m})$ with a collapse threshold of $0.07\text{m}$. The \emph{Diverse Dataset} is configured with a container size of $(1.20\text{m},1.00\text{m},1.70\text{m})$ and a collapse threshold of $0.04\text{m}$. The \emph{Long Board Dataset} employs a container size of $(2.50\text{m},1.20\text{m},1.00\text{m})$ and a collapse threshold of $0.07\text{m}$. All datasets and configuration files are available on our project page: \url{https://robot-bin-packing-benchmark.github.io/documentation.html}.

\subsection{Test Setting}
We define three test settings to evaluate algorithm adaptability under increasing levels of physical realism (see Figure~\ref{fig:test setting}).

\textbf{Math Pack:} This setting performs purely geometric placement without physics or robot operation, allowing evaluation to focus solely on the algorithm’s spatial reasoning ability. By removing physical factors such as gravity, friction, and motion uncertainty, it reflects the algorithm’s idealized noise-free upper bound of packing performance.

\textbf{Physics Pack:} In this setting, physical effects such as gravity and collisions are enabled, but no robot is involved. It evaluates whether the placement strategy can withstand real-world physical constraints such as stacking stability, balance, and collision dynamics. This highlights the robustness of the algorithm under realistic conditions, without introducing additional complexities from motion planning or robot execution.

\textbf{Execution Pack:} This is the most realistic evaluation setting, integrating physics simulation and robotic execution, including motion planning and control. Performance in this setting depends on kinematic reachability, collision-free trajectory planning, and stability during execution. This end-to-end evaluation examines the full workflow—from placement planning to executable robot actions—and reflects readiness for real-world deployment.

\subsection{Metric}

\subsubsection{Details}

We summarize previously used metrics, which mainly focus on compactness, stability, and inference efficiency. With the introduction of our physics-based simulation environment, we can assess execution efficiency, which motivates two additional metrics: Collapsed Placement and Dangerous Operation. Table~\ref{tab:metrics_summary} summarizes 8 metrics used in a comprehensive evaluation of algorithm performance.

\textbf{Space Utilization.} The ratio of the total volume of successfully placed boxes (without overlapping or out-of-bounds errors) to the volume of its container.

\textbf{Occupancy.} Equation~\eqref{eq:occupancy} defines it as the ratio between the total volume of all placed objects and the total volume of the occupied heightmap~\citep{ananno2024multi}:
\begin{equation}
\text{Occupancy} = \frac{\sum_{i=1}^{n} v_i}{v_{\mathrm{h}}}
\label{eq:occupancy}
\end{equation}
where $v_i$ denotes the volume of the $i$-th object, and $v_{\mathrm{h}}$ represents the volume implied by the occupied heightmap.
Higher values indicate denser utilization of the occupied region with fewer voids, while lower values imply wasted space—suggesting that the algorithm produces feasible placements but lacks fine-grained packing efficiency.

\textbf{Decision Time.} Equation~\eqref{eq:decision_time} defines the mean decision-making time per placement:
\begin{equation}
\text{Decision Time} = \frac{\sum_{i=1}^{n} \left( t_{\mathrm{end},i}^{\mathrm{a}} - t_{\mathrm{start},i}^{\mathrm{a}} \right)}{n}
\label{eq:decision_time}
\end{equation}
where $t_{\mathrm{start},i}^{\mathrm{a}}$ and $t_{\mathrm{end},i}^{\mathrm{a}}$ denote the start and end timestamps of the algorithm computation for the $i$-th placement.
Lower values indicate faster decision-making and higher computational efficiency.

\textbf{Local Stability.} Equation~\eqref{eq:position_offset} defines the average Euclidean distance between target and actual placement positions:
\begin{equation}
\text{Local Stability} = \frac{\sum_{i=1}^{n} d_i}{n}
\label{eq:position_offset}
\end{equation}
where $d_i$ is the Euclidean distance between the planned and actual position of the $i$-th object.
Lower values indicate higher placement accuracy and stability, while larger offsets imply that objects slide or tilt away from their intended positions.

Specifically, we record the target position of each box during placement and, after completion, obtain the actual box positions from PyBullet to compute Local Stability. If the calculated Local Stability exceeds a predefined scene threshold, the scene is regarded as collapsed.

\begin{table*}[t]
\centering
\caption{Summary of metrics used in online bin packing. The first column indicates the type of metric. $^\ast$ indicates the two metrics that we propose: Collapsed Placement and Dangerous Operation.}
\scalebox{0.9}{
\begin{tabular}{l |l | c |l}
\hline
\textbf{Type} & \textbf{Metric} &  \textbf{Weight}  & \textbf{Description} \\
\hline
\multirow{2}{*}{Compactness} 
 & Space Utilization & 0.35  & Ratio placed item volume to the container volume. \\
 & Occupancy & 0.15  & Ratio of placed item volume to occupied space. \\
\hline
Inference Efficiency & Decision Time & 0.08  & Average algorithmic decision time per placement. \\
\hline
\multirow{2}{*}{Stability} 
 & Local Stability & 0.07  & Average distance between target and actual positions. \\
 & Static Stability & 0.15  & Rewards based on the linear and angular velocities of the box. \\
\hline
\multirow{3}{*}{Execution Efficiency} 
 & Trajectory Length & 0.08  & Average end-effector trajectory length. \\
 & Collapsed Placement$^\ast$ & 0.07  & Fraction of placements that collapse.  \\
 & Dangerous Operation$^\ast$ & 0.05  & Fraction of unsafe robotic actions. \\
\hline
\end{tabular}
}
\label{tab:metrics_summary}
\end{table*}

\textbf{Static Stability.} Quantifies the overall resistance of the stacked structure to small perturbations, adapted from \cite{icra2023}.
\wzf{It measures whether the arrangement remains stable under small oscillatory base motions, which correspond to the residual motions of objects after placement in the PyBullet simulation caused by slight imbalance and captured through several additional simulation steps. }After each placement, we record linear and angular velocities over 200 PyBullet simulation steps.
For each object $i$:
\begin{equation}
v_{\mathrm{max},i} = \max_t \bigl( v(i,t) \bigr)
\end{equation}
and the mean velocity across all $n$ objects is:
\begin{equation}
\bar{v} = \frac{1}{n} \sum_{i=1}^{n} v_{\mathrm{max},i}
\end{equation}

The linear and angular stability rewards are computed as:

\begin{equation}
R_{\mathrm{lin}} = \max(0, \min(1, -(\bar{v}_{\mathrm{lin}})^{0.4} + 1))
\end{equation}

\begin{equation}
R_{\mathrm{ang}} = \max(0, \min(1, -(\bar{v}_{\mathrm{ang}})^{0.3} + 1))
\end{equation}
and the final stability reward:
\begin{equation}
R = 0.5 R_{\mathrm{lin}} + 0.5 R_{\mathrm{ang}}
\label{eq:static_stability}
\end{equation}
This metric complements Local Stability by evaluating global stack stability instead of local placement accuracy.

In fact, we run 200 simulation steps and record the linear and angular velocities of all boxes. For each box, we take the maximum measured velocity as its representative value, then select the overall maximum among all boxes. Finally, the linear and angular stability rewards are calculated from these maximum values and averaged to obtain the final stability reward.

\textbf{Trajectory Length.} Equation~\eqref{eq:trajectory_length} defines the mean end-effector trajectory length per placement \citep{shuai2023compliant}:
\begin{equation}
\text{Trajectory Length} = \frac{\sum_{i=1}^{n} l_i}{n}
\label{eq:trajectory_length}
\end{equation}
where $l_i$ is the end-effector trajectory length for the $i$-th placement.
Shorter paths indicate easier placements with more robot-friendly trajectories. 
Because the robot avoids collisions with already placed boxes, suboptimal placements often lead to longer paths.
This metric should be interpreted jointly with Space Utilization, as infeasible runs may trivially reduce trajectory length.

\textbf{Collapsed Placement.} Equation~\eqref{eq:collapsed_placement} defines the proportion of placements that collapse during the simulation:
\begin{equation}
\text{Collapsed Placement} = \frac{n_{\mathrm{c}}}{n}
\label{eq:collapsed_placement}
\end{equation}
where $n_{\mathrm{c}}$ is the number of collapsed placements, and $n$ is the total number of feasible placements.
Lower values indicate that the algorithm achieves both geometric feasibility and physical robustness, ensuring that the placed objects remain stable throughout the process.

\textbf{Dangerous Operation.} Equation~\eqref{eq:dangerous_operation} defines the proportion of robotic grasping actions exceeding a predefined safe execution-time threshold:
\begin{equation}
\text{Dangerous Operation} = \frac{n_{\mathrm{d}}}{n}
\label{eq:dangerous_operation}
\end{equation}
where $n_{\mathrm{d}}$ is the number of grasping actions that exceed the predefined safety threshold, and $n$ is the total number of grasping actions.
Lower values indicate that the algorithm generates more robot-friendly placements, allowing safe and efficient manipulation.
Combined with Trajectory Length, it reflects motion efficiency; combined with Space Utilization, it indicates whether feasible placements are also safe for robotic execution.

\subsubsection{Scoring System}

To obtain an overall evaluation of algorithmic performance across multiple metrics, we design a scoring system that converts all metrics into unified normalized scores and computes a weighted aggregate score \citep{aydin2020multi}. 

Since the original metrics differ in scale and optimization direction (some are “higher-is-better,” while others are “lower-is-better”), we first normalize all metrics into the range $[0,1]$ using min–max normalization. For a metric $x$, the normalized score $s(x)$ is computed as:

\begin{equation}
s(x) = 
\begin{cases}
\dfrac{x - x_{\min}}{x_{\max} - x_{\min}}, & \text{if higher is better,} \\
\\[-0.5em]
\dfrac{x_{\max} - x}{x_{\max} - x_{\min}}, & \text{if lower is better.}
\end{cases}
\label{eq:normalization}
\end{equation}

This transformation ensures that higher scores consistently represent better performance across all metrics.  
If a metric has identical values across all algorithms (i.e., $x_{\max} = x_{\min}$), we assign $s(x) = 0$, indicating no discriminative contribution.

After normalization, the overall performance score of each algorithm is computed as a weighted sum of all normalized metrics:

\begin{equation}
\text{Score} = \sum_{j=1}^{m} w_j \, s_j, \qquad \sum_{j=1}^{m} w_j = 1,
\label{eq:weighted_score}
\end{equation}

where $s_j$ denotes the normalized score of the $j$-th metric and $w_j$ is its corresponding weight.  
The detailed weight assignment for each metric is summarized in Table~\ref{tab:metrics_summary}.

The weight vector in the Execution Pack is defined as
\[
w_{\text{exec}} = (0.35,\, 0.15,\, 0.08,\, 0.07,\, 0.15,\, 0.08,\, 0.07,\, 0.05),
\]

The above weight distribution reflects a balanced yet goal-oriented evaluation philosophy:
\textbf{Space Utilization (0.35)} receives the largest weight since it directly measures the effective use of container volume—the primary objective in industrial packing tasks.
\textbf{Occupancy (0.15)} and \textbf{Static Stability (0.15)} are assigned moderate weights, as they respectively characterize fine-grained packing compactness and global structural robustness.
\textbf{Decision Time (0.08)} and \textbf{Trajectory Length (0.08)} capture computational and motion efficiency, which are essential for real-time robotic deployment.
\textbf{Local Stability (0.07)}, \textbf{Collapsed Placement (0.07)}, and \textbf{Dangerous Operation (0.05)} evaluate execution feasibility and safety of manipulation. Their inclusion ensures that algorithms achieving high utilization without sufficient robustness or safety will be penalized.

In the Execution Pack, 8 evaluation metrics are available for measurement.  
In contrast, the Physics Pack only measures 5 metrics, excluding Trajectory Length, Collapsed Placement, and Dangerous Operation. Accordingly, the weight vector is adjusted and normalized as
\[
w_{\text{phys}} = (0.43,\, 0.19,\, 0.10,\, 0.09,\, 0.19).
\]
Similarly, the Math Pack evaluates only three metrics, omitting Local Stability and Static Stability, 
and the corresponding weight vector is given by
\[
w_{\text{math}} = (0.60,\, 0.26,\, 0.14).
\]
These adjusted weightings ensure consistent relative importance among retained metrics while maintaining $\qquad \sum_{j=1}^{m} w_j = 1$.

Overall, this scoring system emphasizes efficient space utilization while maintaining stability, safety, and execution feasibility—yielding a fair, holistic score for each algorithm.

\begin{table*}[t]
\centering
\caption{\textbf{Overall Performance Ranking.} The top-4 results are highlighted as \colorbox{colorfirst}{~first~},
\colorbox{colorsecond}{~second~},
\colorbox{colorthird}{~third~},
and \colorbox{colorfourth}{~fourth~}. Algorithm categories are denoted by GRL (Geometry-aware RL Methods), RRL (Rule-guided RL Methods), RGH (Rule-based Greedy Heuristics), and EDH (EMS-driven Heuristics).}
\scalebox{0.85}{
\begin{tabular}{c|c|ccc|ccc|ccc}
\hline
\multicolumn{2}{c|}{Setting} & \multicolumn{3}{c|}{Math Pack} & \multicolumn{3}{c|}{Physics Pack} & \multicolumn{3}{c}{Execution Pack} \\
\hline
\multicolumn{2}{c|}{Dataset} & Repetitive & Diverse & Long Board & Repetitive & Diverse & Long Board & Repetitive & Diverse & Long Board \\
\hline
\multirow{3}{*}{GRL} 
& PCT        & \cellsecond{0.908}  & \cellfirst{0.939}  & \cellfirst{0.971}  & 0.810  & \cellfirst{0.714}  & \cellfirst{0.839}  & \cellthird{0.765}  & \cellfourth{0.672}  & \cellfirst{\wzf{0.809}} \\
& TAP-Net++  & 0.797  & 0.387  & \cellfourth{0.815}  & \cellfirst{0.891}  & 0.502  & \cellthird{0.725}  & \cellsecond{0.781}  & 0.603  & \cellthird{\wzf{0.676}} \\
& AR2L       & \cellfirst{0.914}  & \cellsecond{0.878}  & \cellthird{0.824}  & 0.740  & \cellthird{0.611}  & \cellfourth{0.694}  & \cellfourth{0.708}  & \cellfirst{0.737}  & \wzf{0.667} \\
\hline
\multirow{2}{*}{RRL} 
& PackE      & 0.397  & 0.440  & 0.207  & \cellsecond{0.865}  & \cellsecond{0.623}  & 0.209  & 0.617  & 0.469  & \wzf{0.272} \\
& CDRL       & 0.468  & 0.638  & 0.655  & 0.491  & 0.453  & 0.612  & 0.422  & \cellsecond{0.728}  & \cellfourth{\wzf{0.672}} \\
\hline
\multirow{4}{*}{RGH} 
& DBL        & \cellfourth{0.854}  & \cellfourth{0.763}  & \cellsecond{0.879}  & \cellfourth{0.850}  & 0.476  & \cellsecond{0.816}  & \cellfirst{0.807}  & \cellthird{0.713}  & \cellsecond{\wzf{0.802}} \\
& LSAH       & \cellthird{0.863}  & \cellthird{0.794}  & 0.679  & \cellthird{0.861}  & 0.477  & 0.510  & 0.701  & 0.523  & \wzf{0.441} \\
& HM         & 0.823  & 0.627  & 0.609  & 0.709  & \cellfourth{0.554}  & 0.683  & 0.414  & 0.623  & \wzf{0.670} \\
& SDFPack    & 0.659  & 0.579  & 0.087  & 0.389  & 0.345  & 0.193  & 0.232  & 0.228  & \wzf{0.174} \\
\hline
\multirow{3}{*}{EDH} 
& OnlineBPH  & 0.527  & 0.534  & 0.583  & 0.517  & 0.478  & 0.560  & 0.464  & 0.618  & \wzf{0.623} \\
& MACS       & 0.221  & 0.135  & 0.009  & 0.298  & 0.183  & 0.236  & 0.238  & 0.364  & \wzf{0.216} \\
& BR         & 0.781  & 0.677  & 0.396  & 0.542  & 0.330  & 0.291  & 0.410  & 0.360  & \wzf{0.274} \\
\hline
\end{tabular}
}
\label{tab:overall}
\end{table*}

\section{Evaluation}
\wzf{Our experiments aim to answer the following questions about the tested algorithms}: (1) Which categories of algorithms are more suitable for the three industrial scenarios we propose? (2) Do the algorithms exhibit consistent performance patterns across different scenarios or test settings? (3) Which algorithms excel in particular evaluation dimensions (metrics)? (4) Can the stability-related metrics from the simulation environment guide the model toward producing stable and feasible results? \wzf{(5) How do different algorithms compare in terms of the estimated industrial throughput measured by theoretical Picks Per Hour (PPH)? (6) To what extent do the simulation results align with the behavior observed on a real robotic system?} To answer these questions, we conduct several groups of experiments.

\subsection{Overall Performance Analysis}
We reproduced the closed-source algorithms based on the descriptions provided in their original papers, with implementation details presented in the appendix. All learning-based methods were trained on a unified training set, and the proposed datasets were divided into training and testing subsets. Each dataset adopted a sampling strategy consistent with its characteristics: consecutive sequences for the Repetitive Dataset, proportion-based sampling for the Diverse Dataset, and random sampling for the Long Board Dataset. For evaluation, 30 test groups were drawn from a private testing pool following the same principles to ensure fairness and mitigate randomness. The container dimensions were consistent with those in real industrial scenarios. In the simulation environment, packing terminates upon a collapse event, a placement deviation exceeding the threshold, or a box being placed beyond the container height. Unified testing was conducted under three settings and three datasets, leading to several key findings.

We aggregated the results of the above experiments using the scoring system to compute an overall score for each algorithm, which allows us to rank all methods across different test settings and datasets, as shown in Table~\ref{tab:overall}. Since the Execution Pack is the most important setting, as it provides the closest reflection of real industrial deployment, we first analyze which algorithms perform best under this scenario.

\subsubsection{Analysis on the Repetitive Dataset}

On the Repetitive Dataset under the Execution Pack setting, all three geometry-aware RL algorithms (PCT, TAP-Net++, and AR2L) rank within the top 4 overall, as shown in Table~\ref{tab:overall}. The key characteristic of this dataset is its high level of repetition in box types. The ability to capture these recurring spatial patterns and generalize across sequences is crucial for strong performance on this dataset~\citep{pajarinen2017robotic,holladay2024leveraging}.

In particular, PCT demonstrates strong performance in both Compactness and Execution Efficiency. It stems from two factors: the graph attention network (GAT) models spatial configurations and geometric dependencies among repetitive boxes, enabling stable and compact arrangements. Meanwhile, the heuristic-guided leaf node expansion (e.g., EMS/EV schemes) efficiently generates high-potential placements for homogeneous geometries. TAP-Net++ exhibits good Compactness and strong Stability. Its EMS-based geometric matching is well-suited to repetitive-box scenarios, confirming that efficient space allocation helps preserve stable sequential layouts. AR2L also performs well on the Repetitive Dataset because its permutation-based attacker plays a key role in strengthening the model’s ability to capture spatial and geometric relationships. By continually exposing the policy to challenging permutations of highly similar boxes, AR2L is encouraged to rely on intrinsic geometric relations~\citep{hsu2023sim,jiang2024transic}. 

Overall, these results highlight that \textbf{RL algorithms capable of modeling spatial and geometric relationships perform well on the Repetitive Dataset.}

\subsubsection{Analysis on the Diverse Dataset}

On the Diverse Dataset under the Execution Pack setting, PCT and  AR2L rank within the top 4 overall, as shown in Table~\ref{tab:overall}. In the Diverse Datasets, box sizes vary widely without a clear pattern. Heuristic methods rely on fixed rules (e.g., largest area) and lack global adaptability~\citep{bortfeldt2013constraints}, so they may struggle to efficiently place boxes when size differences are large. In contrast, RL methods learn adaptive placement policies that consider the current box and the overall packing state. They can dynamically adjust placements, implicitly capture spatial interactions, and better handle large size variations. Therefore, RL generally performs better than heuristics in such scenarios.

Specifically, PCT maintains leading performance in Geometric Compactness and Stacking Stability, while also performing well in Physical Feasibility. Its Transformer backbone with hierarchical representations models geometric compatibilities between item classes, and the global spatial-relational reasoning enables structural generalization. This allows PCT to effectively learn the dimensional variability of the Diverse Dataset, extending its strengths to heterogeneous industrial layouts. AR2L adopts the same Transformer-based network as its packing policy and employs adversarial training to mitigate inter-object interference~\citep{pinto2017robust}. Its permutation-based attacker naturally leverages the size difference between boxes to generate worst-case sequences, helping the policy learn more robust models. Although TAP-Net++ also incorporates a Transformer-based packing policy, its internal components limit the effectiveness of this module, leading to only moderate performance on the Diverse Dataset~\citep{mishra2023convolutional}. This counterintuitive behavior will be examined in detail in the Cross Comparison section.

The Transformer-based packing policy can better model the size and positional features of already placed boxes — for instance, understanding which critical regions are already occupied by large boxes and where smaller boxes can be appropriately placed. Overall, these results highlight that \textbf{RL methods with a Transformer-based packing policy, which can better learn the distribution of item sizes, perform well on the Diverse Dataset.}

\subsubsection{Analysis on the Long Board Dataset}
On the Long Board Dataset under the Execution Pack setting, PCT, TAP-Net++, and  DBL rank within the top 3 overall, as shown in Table~\ref{tab:overall}. EMS-based RL methods explicitly leverage geometric constraints and perform matching over feasible spaces, which is particularly advantageous for elongated items requiring precise placement. By discretizing the search space into stable candidate regions and enabling fine-grained alignment between boxes and feasible spaces, these methods effectively handle the extreme structural characteristics and tight spatial feasibility of the Long Board Dataset. In this dataset, feasible regions are extremely narrow, with items exhibiting nearly fixed orientations predominantly aligned along the main axis~\citep{bennell2008geometry}. In such a geometrically constrained environment, single-rule greedy heuristics naturally conform to deterministic spatial constraints, even when using simple placement rules such as placing at the lowest height or aligning to the left boundary. Furthermore, their single-rule designs directly support the core objectives of Geometric Compactness, Stacking Stability, and Physical Feasibility, enabling compact and stable layouts without relying on complex search or learning procedures.

Specifically, TAP-Net++ combines EMS-based placement with joint encoding and feasibility-aware matching, leveraging geometric constraints to identify and exploit the elongated feasible regions typical of this dataset, resulting in highly compact and stable layouts. Similarly, PCT performs well on the Long Board Dataset due to its tree-structured representation and EMS-based placement. The hierarchical representation captures geometric relationships between items, while EMS-based placement identifies feasible candidate positions within these elongated regions, effectively respecting the geometric constraints. Together, these mechanisms produce layouts that are compact, stable, and physically feasible. DBL also performs effectively on the Long Board Dataset by greedily selecting the lowest and most-left feasible positions, thereby promoting stable and compact placements. Its strong geometric prior aligns naturally with the elongated feasible regions, making it effective even without complex learning or search procedures~\citep{crainic2009ts2pack}. 

Overall, these results highlight that \textbf{RL algorithms and heuristics based on geometric constraints perform well on the Long Board Dataset.}

\subsection{Cross Comparison}  
The previous sections explained which algorithms perform best on each dataset. We now examine how individual algorithms behave across different datasets and test settings, highlighting both consistent patterns and cases where performance varies substantially.

\subsubsection{Cross-Dataset Performance Comparison}  

Table~\ref{tab:overall} shows that across all three datasets under the Execution Pack setting, PCT, AR2L, and DBL consistently rank within the top three. Both PCT and AR2L employ Transformer-based packing strategies capable of capturing universal geometric patterns. Their learned spatial reasoning generalizes effectively across repetitive, diverse, and elongated box arrangements, enabling consistently high performance under different dataset characteristics. By contrast, DBL relies on simple heuristic rules that prioritize conservative and contiguous placements. While its design is straightforward, this approach often achieves good space utilization and robust stacking, making it reliable across varied industrial packing tasks. \textbf{PCT, AR2L, and DBL are therefore well-suited for general industrial scenarios.}

\begin{table*}[t]
\centering
\caption{\textbf{Occupancy Performance Across Test Settings and Datasets.} Each entry reports the Occupancy achieved by an algorithm under three test settings and three dataset types. The comparison is performed column-wise. The best and second-best
results in each column are highlighted using \colorbox{colorfirst}{~first~} and \colorbox{colorsecond}{~second~}, respectively.}
\scalebox{0.9}{
\begin{tabular}{c|ccc|ccc|ccc}
\hline
Setting & \multicolumn{3}{c|}{Math Pack} & \multicolumn{3}{c|}{Physics Pack} & \multicolumn{3}{c}{Execution Pack} \\
\hline
Dataset & Repetitive & Diverse & Long Board & Repetitive & Diverse & Long Board & Repetitive & Diverse & Long Board \\
\hline
PCT        & 0.895  & 0.897  & \cellsecond{0.771}  & 0.942  & 0.921  & 0.802  & 0.944  & 0.964  & 0.818 \\
TAP-Net++  & 0.915  & \cellsecond{0.936}  & 0.724  & 0.948  & 0.962  & 0.801  & 0.952  & 0.976  & 0.812 \\
AR2L       & 0.898  & 0.878  & 0.716  & 0.958  & 0.867  & 0.792  & 0.959  & 0.918  & 0.809 \\
\hline
PackE      & \cellfirst{0.965}  & 0.898  & 0.647  & \cellfirst{0.965}  & \cellfirst{0.986}  & 0.687  & \cellsecond{0.965}  & 0.918  & 0.699 \\
CDRL       & 0.826  & 0.834  & 0.699  & 0.888  & 0.871  & 0.759  & 0.927  & 0.931  & 0.774 \\
\hline
DBL        & 0.911  & 0.886  & 0.769  & 0.947  & 0.956  & \cellsecond{0.815}  & 0.948  & 0.968  & \cellsecond{0.826} \\
LSAH       & 0.908  & 0.885  & 0.686  & 0.943  & 0.947  & 0.748  & 0.945  & 0.967  & 0.778 \\
HM         & \cellsecond{0.960}  & \cellfirst{0.943}  & \cellfirst{0.787}  & \cellfirst{0.965}  & 0.965  & \cellfirst{0.831}  & \cellfirst{0.966}  & \cellsecond{0.980}  & \cellfirst{0.849} \\
SDFPack    & 0.834  & 0.849  & 0.619  & 0.918  & 0.909  & 0.732  & 0.929  & 0.94   & 0.761 \\
\hline
OnlineBPH  & 0.739  & 0.768  & 0.676  & 0.941  & \cellsecond{0.975}  & 0.745  & 0.940  & \cellfirst{0.981}  & 0.758 \\
MACS       & 0.648  & 0.687  & 0.610  & 0.911  & 0.880  & 0.738  & 0.915  & 0.885  & 0.75 \\
BR         & 0.873  & 0.836  & 0.627  & 0.927  & 0.895  & 0.729  & 0.942  & 0.942  & 0.775 \\
\hline
\end{tabular}
}
\label{tab:single-occupancy}
\end{table*}

TAP-Net++ performs well on the Repetitive and Long Board datasets but only moderately on the Diverse dataset under the Execution Pack setting (the ranking results are presented in Table~\ref{tab:overall}). Its stability-focused design, including the Cross-Modal Transformer, EMS-based geometric matching, and feasibility-guided RL, enables robust and precise placements in structured or elongated box arrangements. In Repetitive data, repeated patterns allow stable stacking, and in Long Board data, EMS-based geometric matching ensures accurate handling of elongated items. However, the Diverse dataset’s highly heterogeneous box sizes limit the effectiveness of its stability-focused strategy and action pruning, leading to overly conservative placements that may overlook more flexible or innovative positioning opportunities, which further reduces space utilization. Overall, \textbf{TAP-Net++ performs well when the dataset structure aligns with its stability-oriented design but becomes less adaptable in highly variable scenarios.}

\subsubsection{Cross-Setting Performance Comparison}  
In this section, we analyze how the rankings of the top-4 algorithms vary across the different test settings.

For the Repetitive Dataset, the top-4 algorithms remain largely stable across different settings. This dataset emphasizes continuous stability, and small errors are easily amplified in both the Physics Pack and Execution Pack. Therefore, methods that are structurally robust and choose positions feasible for robotic execution, rather than purely geometrically optimal, tend to perform better. As a result, approaches like DBL and TAP-Net++ surpass AR2L and PCT under these settings. DBL’s strategy rarely obstructs future placements, allowing its strong performance in Math Pack to translate effectively to Physics Pack and Execution Pack. TAP-Net++, on the other hand, inherently selects positions guided by stability, favoring safer and more conservative placements. While its Math Pack performance is moderate, it overtakes others in the more physically constrained settings.

In the Diverse dataset, algorithm rankings fluctuate noticeably. The large variation in box sizes leads to complex physical interactions, requiring algorithms to possess strong adaptability and self-adjustment capabilities. Consequently, RL-based methods outperform heuristic approaches under realistic execution settings. Many heuristics are designed to optimize stacking density under the assumption of perfect placement and ignore physical factors, relying on precise fitting between differently sized boxes. When physical effects are considered, heterogeneous sizes increase stacking risk: small boxes may wedge unpredictably or block space for future larger boxes, while large boxes placed on smaller supports may topple. In contrast, learning-based methods account for stability and robot-friendly placement, which is particularly effective for datasets with wide size variations.

For the Long Board Dataset, the top-performing algorithms remain nearly unchanged across all settings. This stability suggests that tasks dominated by geometric constraints—such as elongated or panel-like shapes—are less sensitive to physical or execution-level complexities. In these cases, precise geometric control and balanced placement remain the dominant performance factors.

\begin{table}[t]
\centering
\caption{\textbf{Trajectory Length Performance on Execution Pack.} Each entry reports the Trajectory Length (lower is better) under three datasets. The comparison is conducted column-wise.}
\scalebox{0.9}{
\begin{tabular}{c|ccc}
\hline
Dataset & Repetitive & Diverse & Long Board \\
\hline
PCT          & 2.380  & \cellfirst{1.989}  & 3.053 \\
TAP-Net++    & 2.371  & 2.115  & 3.127 \\
AR2L         & 2.351  & \cellsecond{2.109}  & \cellsecond{2.601} \\
\hline
PackE        & \cellsecond{2.200}  & 2.495  & \cellfirst{1.916} \\
CDRL         & \cellfirst{2.104}  & 2.347  & 2.916 \\
\hline
DBL          & 2.253  & 2.206  & 3.226 \\
LSAH         & 2.313  & 2.231  & 3.095 \\
HM           & 2.475  & 2.221  & 3.361 \\
SDFPack      & 2.306  & 2.148  & 3.593 \\
\hline
OnlineBPH    & 2.790  & 2.232  & 2.987 \\
MACS         & 2.744  & 2.378  & 3.808 \\
BR           & 2.454  & 2.212  & 3.642 \\
\hline
\end{tabular}
}
\label{tab:single-trajectory}
\end{table}

\subsection{Single Metric Analysis}  
While overall performance rankings provide a useful high-level summary, they inevitably hide important differences in how algorithms behave on specific evaluation dimensions. Two algorithms with similar overall scores may excel for entirely different reasons. Some metrics correspond directly to core industrial requirements: Occupancy reflects space efficiency, Trajectory Length determines robotic execution cost and feasibility, and Collapsed Placement measures physical stability and safety. By examining these metrics separately, we can reveal performance characteristics that are not captured by the global score and identify which algorithmic design choices drive strengths in particular operational aspects. These strengths provide guidance for algorithm selection in real-world packing tasks. Table~\ref{tab:single-occupancy}, \ref{tab:single-trajectory}, and \ref{tab:single-collapsed} report the results for Occupancy, Trajectory Length, and Collapsed Placement, respectively.

\begin{table}[htbp]
\centering
\caption{\textbf{Collapsed Placement Performance on Execution Pack.} Each entry reports the Collapsed Placement (lower is better) under three datasets. The comparison is conducted column-wise.}

\scalebox{0.9}{
\begin{tabular}{c|ccc}
\hline
Dataset & Repetitive & Diverse & Long Board \\
\hline
PCT          & \cellfirst{0.033}  & 0.452  & 0.139 \\
TAP-Net++    & 0.068  & 0.544  & \cellfirst{0.113} \\
AR2L         & 0.073  & 0.538  & 0.188 \\
\hline
PackE        & 0.302  & \cellsecond{0.408}  & 0.22  \\
CDRL         & 0.676  & \cellfirst{0.256}  & 0.135 \\
\hline
DBL          & \cellsecond{0.041}  & 0.478  & \cellsecond{0.119} \\
LSAH         & 0.108  & 0.652  & 0.245 \\
HM           & 0.709  & 0.651  & 0.17  \\
SDFPack      & 0.385  & 0.766  & 0.36  \\
\hline
OnlineBPH    & 0.157  & 0.552  & 0.151 \\
MACS         & 0.653  & 0.499  & 0.252 \\
BR           & 0.248  & 0.815  & 0.363 \\
\hline
\end{tabular}
}
\label{tab:single-collapsed}
\end{table}

\begin{table*}[t]
\centering
\caption{\textbf{Results of Stability-related Metrics Exploration Experiment on Repetitive and Diverse Datasets}, tested under Physics Pack. Metric abbreviations: 
Uti. = Space Utilization, 
Occ. = Occupancy, 
Local Stab. = Local Stability, 
Static Stab. = Static Stability. Arrows indicate optimization direction ($\uparrow$: higher better, $\downarrow$: lower better). TAP. = TAP-Net++.}
\scalebox{0.9}{
\setlength{\tabcolsep}{6pt}
\renewcommand{\arraystretch}{1.0}
\begin{tabular}{@{}c|c@{}}
\hline
\multicolumn{1}{c|}{\textbf{Repetitive Dataset}} & \multicolumn{1}{c}{\textbf{Diverse Dataset}} \\
\hline
\begin{tabular}{c|ccccc}
\hline
Model  & \makecell{Uti. $\uparrow$} 
       & \makecell{Occ. $\uparrow$} 
       & \makecell{Local \\ Stab. $\downarrow$} 
       & \makecell{Static \\ Stab. $\uparrow$} \\
\hline
$\mathrm{TAP.}^*$   & \cellfirst{0.402} & \cellfirst{0.941}  & 0.011 & 0.400 \\
TAP.                 & 0.257 & 0.938  & \cellfirst{0.006} & \cellfirst{0.413} \\
\hline
$\mathrm{PCT}^*$     & \cellfirst{0.375} & 0.940 & 0.013 & \cellfirst{0.385} \\
PCT                  & 0.324 & \cellfirst{0.945} & \cellfirst{0.011} & 0.384 \\
\hline
$\mathrm{PackE}^*$   & \cellfirst{0.357} & \cellfirst{0.965} & 0.013 & 0.419 \\
PackE                 & 0.281 & 0.959 & \cellfirst{0.008} & \cellfirst{0.425} \\
\hline
$\mathrm{AR2L}^*$    & 0.201 & 0.892 & 0.012 & 0.400 \\
AR2L                 & \cellfirst{0.367} & \cellfirst{0.942} & \cellfirst{0.009} & \cellfirst{0.406} \\
\hline
\end{tabular}
& 
\begin{tabular}{c|ccccc}
\hline
Model  & \makecell{Uti. $\uparrow$} 
       & \makecell{Occ. $\uparrow$} 
       & \makecell{Local \\ Stab. $\downarrow$} 
       & \makecell{Static \\ Stab. $\uparrow$} \\
\hline
$\mathrm{PCT}^*$   & \cellfirst{0.513} & 0.923 & 0.012 & 0.420 \\
PCT                 & 0.470 & \cellfirst{0.929} & \cellfirst{0.011} & \cellfirst{0.435} \\
\hline
$\mathrm{AR2L}^*$  & 0.395 & 0.856 & \cellfirst{0.014} & \cellfirst{0.440} \\
AR2L                & \cellfirst{0.447} & \cellfirst{0.929} & 0.020 & 0.439 \\
\hline
$\mathrm{TAP.}^*$  & \cellfirst{0.237} & \cellfirst{0.967} & \cellfirst{0.003} & 0.449 \\
TAP.                & 0.232 & 0.950 & 0.004 & \cellfirst{0.454} \\
\hline
$\mathrm{PackE}^*$ & \cellfirst{0.184} & \cellfirst{0.985} & \cellfirst{0.001} & \cellfirst{0.472} \\
PackE               & 0.170 & \cellfirst{0.985} & \cellfirst{0.001} & 0.467 \\
\hline
\end{tabular} \\
\hline
\end{tabular}
}
\label{tab:metric_validity}
\end{table*}

Rule-based Greedy Heuristics, such as HM, naturally fill the lowest available spaces. HM consistently achieves the highest Occupancy (e.g., 0.943 in Diverse Dataset, 0.849 in Long Board Dataset, Occupancy is presented in Table~\ref{tab:single-occupancy}). PackE explicitly selects positions that minimize gaps(Occupancy 0.986 in Diverse Dataset). This compresses the stacking structure and reduces spatial redundancy, leading to improved local stability. Similarly, learning-based methods that incorporate geometry-aware representations and feasible-space matching (e.g., PCT, TAP-Net++) can also produce dense layouts by reasoning about item–space compatibility. \textbf{Algorithms tend to achieve high Occupancy when they prioritize compact and space-efficient placements rather than simply maximizing the number of boxes placed.}

We only consider Trajectory Length under the Execution Pack, and the results are presented in Table~\ref{tab:single-trajectory}. AR2L leverages adjustable robustness training to guide the policy toward stable, low-risk placements, learn to avoid difficult or constrained placement patterns, and employ mixture dynamics to encourage globally smoother geometries (AR2L achieves top-two trajectory lengths on the Diverse and Long Board Datasets). PackE generates compact, contiguous placements that minimize detours (Trajectory Length 1.916 on Long Board Dataset). Overall, any algorithm that emphasizes locally reachable positions, maintains stacking continuity, and reduces collision potential tends to be robot-friendly with short trajectories. \textbf{Algorithms tend to achieve short average end-effector trajectories when their placement strategies prioritize feasible, low-risk positions and produce smooth, structured stacks.}

We only consider Collapsed Placement under the Execution Pack, and the results are presented in Table~\ref{tab:single-collapsed}. On the Long Board Dataset, TAP-Net++ achieves a Collapsed Placement rate of 0.113 through EMS-based placement and feasibility-aware matching. TAP-Net++ chooses placement positions based on stability, resulting in placements that rarely collapse. In most scenarios, DBL maintains the lowest Collapsed Placement by using a layer-by-layer stacking policy, which often results in consecutively placed positions. This placement pattern is generally compact and less prone to collapse, implicitly taking stability into account. These low Collapsed Placements arise because both types of algorithms ensure that each placement is well-supported, avoid collisions, and respect stacking stability. In general, \textbf{algorithms achieve low Collapsed Placement rates when their placement strategies explicitly or implicitly prioritize stability and physical feasibility.}

\subsection{Exploration on Stability-related Metrics}
We evaluated the effectiveness of stability-related metrics in guiding model training. We tested four learning-based methods—TAP-Net++, PCT, PackE, and AR2L—excluding CDRL, which does not incorporate stability estimation in its design. For each method, we created two variants: one trained with reward signals derived from our industrial stability metrics—Static Stability and Local Stability—and another trained without metric-based rewards under idealized geometric conditions, referred to as the baseline model ($\mathrm{model}^*$). We chose these two stability metrics because our simulator provides accurate physical feedback, enabling precise and reliable assessment of stacking stability during training. Thus, they offer realistic supervisory signals that cannot be obtained from purely geometric rewards.

All models were tested under the Physics Pack, which simulates realistic physical interactions without robotic arm execution. We conduct evaluation only in this setting because incorporating robot control would considerably increase RL training cost, making such ablations impractical. In contrast, pure physics simulations allow us to obtain realistic and consistent assessments of stability while keeping computational costs manageable.

Results show that most models trained with metric-guided rewards achieved higher Space Utilization than their baseline counterparts, as shown in Table~\ref{tab:metric_validity}. For example, TAP-Net++ improved from 0.257 to 0.402, PCT from 0.324 to 0.375, and PackE from 0.281 to 0.357. This indicates that these models learned to avoid unstable placements and large positional deviations. Notably, introducing stability metrics as reward signals slightly reduces some stability-related indicators, because the models tend to place more items overall—the denser packing naturally increases stacking difficulty. However, the final physics-based performance still improves, as the gains in feasible placements and overall stacking quality outweigh the minor decline in per-placement stability.

\begin{table*}[t]
\centering
\caption{\textbf{Theoretical Picks Per Hour on Three Datasets.} Each entry reports Decision Time (s), Trajectory Length (m), and theoretical PPH. }
\scalebox{0.9}{
\begin{tabular}{c|ccc|ccc|ccc}
\hline
Dataset & \multicolumn{3}{c|}{Repetitive} & \multicolumn{3}{c|}{Diverse} & \multicolumn{3}{c}{Long Board} \\
\hline
Method & \makecell{Deci. \\ Time $\downarrow$} & \makecell{Trajectory \\ Len. $\downarrow$} & \makecell{PPH $\uparrow$} 
       & \makecell{Deci. \\ Time $\downarrow$} & \makecell{Trajectory \\ Len. $\downarrow$} & \makecell{PPH $\uparrow$} 
       & \makecell{Deci. \\ Time $\downarrow$} & \makecell{Trajectory \\ Len. $\downarrow$} & \makecell{PPH $\uparrow$} \\
\hline
PCT        & 0.131 & 2.380 & 394 & 0.042 & 1.989 & $\textbf{418}$ & 0.029 & 3.196 & 333 \\
TAPNet++   & 0.793 & 2.371 & 372 & 0.553 & 2.115 & 381 & 0.285 & 3.277 & 329 \\
AR2L       & 1.518 & 2.351 & 341 & 1.487 & 2.109 & 355 & 0.798 & 2.753 & 352 \\
PackE      & 0.166 & 2.200 & 394 & 0.498 & 2.495 & 370 & 0.152 & 2.010 & $\textbf{398}$ \\
CDRL       & 0.472 & 2.104 & $\textbf{404}$ & 0.267 & 2.347 & 391 & 0.261 & 3.077 & 336 \\
DBL        & 0.611 & 2.253 & 374 & 0.503 & 2.206 & 382 & 0.722 & 3.195 & 328 \\
LSAH       & 0.004 & 2.313 & 379 & 0.002 & 2.231 & 384 & 0.002 & 3.230 & 326 \\
HM         & 0.495 & 2.475 & 373 & 0.317 & 2.221 & 386 & 1.007 & 3.270 & 324 \\
SDFPack    & 12.672 & 2.306 & 189 & 8.067 & 2.148 & 231 & 14.295 & 3.787 & 164 \\
OnlineBPH  & 0.001 & 2.790 & 368 & 0.002 & 2.232 & 384 & 0.002 & 3.019 & 340 \\
MACS       & 1.244 & 2.744 & 350 & 2.577 & 2.378 & 337 & 4.304 & 3.782 & 237 \\
BR         & 0.004 & 2.454 & 381 & 0.004 & 2.212 & 387 & 0.002 & 3.752 & 301 \\
\hline
\end{tabular}
}
\label{tab:pph}
\end{table*}

In contrast, AR2L exhibits the opposite trend: the metric-guided version performs worse than its baseline. This is largely because AR2L’s reliance on a highly expressive Transformer-based policy that already captures global geometric structure effectively under geometric-only training (AR2L ranks within the top-3 in overall performance across all three datasets under the Math Pack setting, as shown in Table~\ref{tab:overall}). When stability metrics are added as reward signals, the training objective becomes more constrained, biasing the policy toward conservative local adjustments rather than broad spatial exploration. As a result, the model becomes less capable of exploiting its strong geometric reasoning ability, leading to reduced placement volume and lower overall space utilization.

These results collectively suggest that while industrial stability metrics generally guide learning-based methods toward more robust and physically feasible strategies, their effectiveness depends on the algorithm’s intrinsic design. Methods that rely on local feasibility reasoning or heuristic-style decision making benefit more from stability-shaped rewards, whereas globally expressive policies like AR2L may experience reduced flexibility when additional constraints are introduced.

\wzf{
\subsection{Theoretical Picks Per Hour}

In industrial robotic picking systems, Picks Per Hour (PPH) is a standard metric for measuring system throughput. 
To relate algorithmic efficiency to practical industrial performance, we estimate a theoretical PPH by combining the algorithm decision time with the estimated execution time of the robot.

The robot execution time consists of the motion time along the planned trajectory and a fixed idle time for grasping and releasing. 
We set a constant idle time of $T_{\mathrm{idle}} = 2\,\mathrm{s}$ per pick-and-place operation, corresponding approximately to 1\,s for grasping and 1\,s for releasing. This idle time is fixed for operational safety and can therefore be estimated independently of the planning method.

The motion time is estimated from the trajectory length $L_i$ and the measured average speed of the robot:
\begin{equation}
T_{\mathrm{motion},i} = \frac{L_i}{\alpha},
\end{equation}
where $\alpha = 0.34\,\mathrm{m/s}$ is the average speed of the real robot, including acceleration and deceleration effects.

The total cycle time for one pick-and-place operation is then
\begin{equation}
T_{\mathrm{cycle},i} = T_{\mathrm{decision},i} + T_{\mathrm{motion},i} + T_{\mathrm{idle}},
\end{equation}
and the corresponding theoretical throughput is
\begin{equation}
PPH_i = \frac{3600}{T_{\mathrm{cycle},i}} = \frac{3600}{T_{\mathrm{decision},i} + L_i/\alpha + T_{\mathrm{idle}}}.
\end{equation}

This formulation contextualizes Decision Time within an estimated industrial cycle time and provides a practical estimate of potential throughput for different packing strategies. 
The resulting theoretical PPH values for all methods across the three datasets are reported in Table~\ref{tab:pph}.

\begin{figure*}[!t]
    \centering
    \includegraphics[width=0.83\textwidth,keepaspectratio]{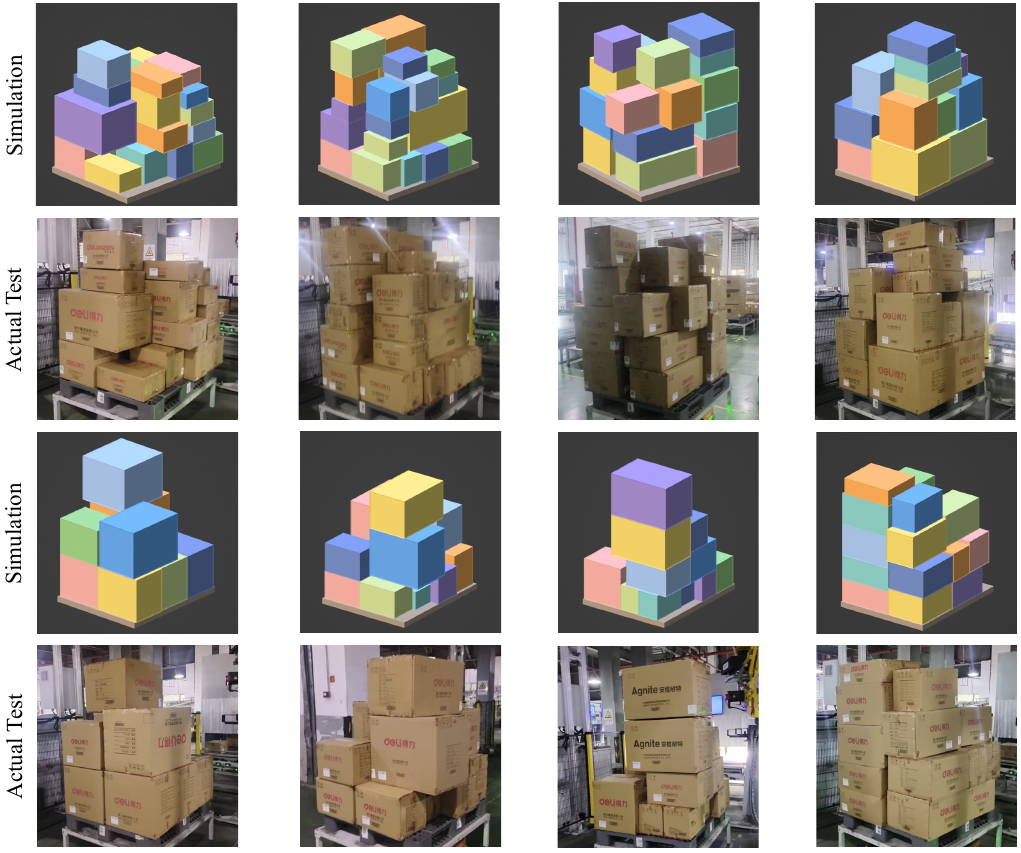}
    \caption{Comparison of the results between simulation and real-world experiments for tests from the Diverse Dataset.}
    \label{fig:real}
\end{figure*}

\subsection{Hardware Testing}

To assess the fidelity of the RoboBPP simulation environment, we conducted real-world validation experiments in collaboration with an industrial partner using  an ABB IRB 6700 robotic arm equipped with a suction-based modular gripper.

In this setup, boxes were delivered by a conveyor belt and detected by a photoelectric sensor. An RGB-D camera (PhoXi 3D Scanner XL) captured the top surface of incoming boxes, and box detection and segmentation were performed using Mask R-CNN. The detected box poses were transformed into the robot coordinate system through eye-to-hand calibration for grasping. The pallet size was $120 \times 100$\,cm, and the maximum stack height was 170\,cm.

Because these experiments were conducted in the warehouse of a real office supplies manufacturer Deli, the deployed tests had to follow the factory's operational constraints. In particular, we were only allowed to validate the PCT method~\citep{zhao2021learning} that had already been delivered to the industrial partner, rather than testing all benchmark methods on the real robot. The results showed strong consistency with simulation: boxes that could be stably placed in the RoboBPP simulation were also stably placed by the real robot.

Figure~\ref{fig:real} compares representative real-world placement results with their simulation counterparts, illustrating this close agreement. These results demonstrate that RoboBPP provides a reasonably faithful approximation of real industrial robotic packing operations, highlighting its suitability for evaluating algorithmic performance under both geometric and physically realistic conditions.

}
\section{Conclusion}
In this work, we introduce RoboBPP, the first comprehensive benchmarking system for robotic online 3D bin packing. Unlike prior studies that treat 3D-BPP purely as a mathematical optimization problem, RoboBPP continuously monitors physical stability, simulates realistic robotic constraints, and incorporates real-world production data, providing more reliable and actionable evaluations. Specifically, RoboBPP includes three datasets covering repetitive, diverse, and long-board scenarios, and three test settings ranging from purely geometric to physics-constrained and full robotic execution. A multi-dimensional metric system with a weighted scoring scheme enables holistic comparison across compactness, stability, inference efficiency, and operational safety. Using this framework, we reproduce a wide range of heuristic and learning-based algorithms and provide insights into their strengths, weaknesses, and scenario-specific applicability. Cross-dataset and single-metric analyses further reveal the algorithm's generalization capabilities and robustness. All resources are publicly available, supporting reproducibility and community collaboration.

\textbf{Limitations and future work.} Despite its contributions, RoboBPP has several inherent limitations. The simulator cannot fully capture all complexities of real-world environments, and currently, all items are modeled as rectangular rigid bodies, limiting applicability to more irregular or deformable objects. Future work will address these limitations by supporting irregular and deformable items, enabling multi-robot collaboration, and validating algorithms on real robotic systems, thereby improving both realism and practical applicability.
\nocite{*}
\bibliographystyle{SageH}
\bibliography{references}


\section{APPENDIX}

\subsection{MORE EXPERIMENTAL DETAILS}

\subsubsection{Dataset DETAILS}

\begin{figure*}[!htp]
    \centering
    \includegraphics[width=\textwidth]{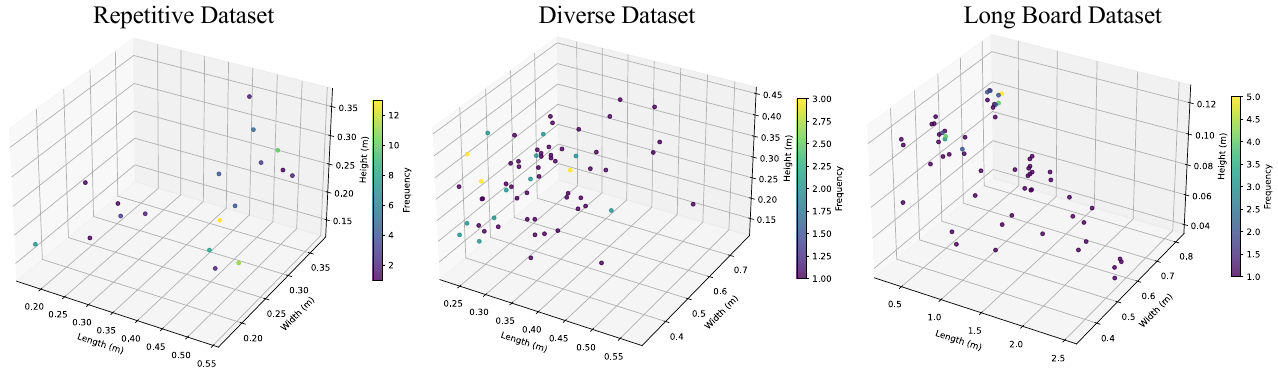}
    \caption{3D scatter plots based on box length, width, and height.}
    \label{fig:scatter}
\end{figure*}

As shown in Figure~\ref{fig:scatter}, we visualize a test data group as 3D scatter plots based on box length, width, and height, with each group containing 80 boxes. The color intensity of each point represents its frequency, enabling us to observe the distinctive structural characteristics of the three datasets. The Repetitive Dataset contains relatively few unique box types after deduplication, and these types occur with high frequency, reflecting the repetitive nature of assembly-line production. The Diverse Dataset features a wide variety of box sizes, consistent with logistics scenarios where item heterogeneity is high. The Long Board Dataset exhibits a much broader distribution along the length axis compared with the other two datasets, highlighting the extreme aspect ratios typical of elongated furniture items.

\subsubsection{Exploration on Stability-related Metrics DETAILS}
We use the PyBullet simulation environment. During training, the metrics \emph{Static Stability} and \emph{Local Stability} are incorporated as reward signals, as shown in Algorithm~\ref{alg:reward_metrics}.

\begin{algorithm}
\caption{Reward with Stability and Local Stability}
\label{alg:reward_metrics}
\begin{algorithmic}[1]
\STATE \textbf{Input:} Current item $b$, placed items $P$, container $C$
\STATE \textbf{Output:} Reward value $r$

\STATE $r \gets box\_ratio \times 10$
\STATE Reset simulation and place all boxes ($P$ and $b$)
\STATE Run simulation for a fixed number of steps

\STATE \textbf{Compute Stability Reward:}
\STATE For each box, record maximum linear and angular velocities
\STATE Convert them into stability scores using formulas in Section 3.4 Metric
\STATE $stability\_reward \gets$ average of stability scores

\STATE \textbf{Compute Local Stability:}
\STATE For each box, compare the target pose and the final pose
\STATE Compute offset along $x$ and $y$ axes
\STATE Take mean value over all boxes as $positions\_offset$

\STATE Update reward:
\[
r \gets r + 0.01 \times stability\_reward - positions\_offset
\]

\STATE \textbf{Return:} $r$
\end{algorithmic}
\end{algorithm}

\subsection{ALGORITHM DETAILS}

In this appendix, we provide pseudocode for all 12 benchmarked algorithms, summarizing their main steps, inputs/outputs, and key computations while omitting low-level details for clarity. This allows a concise comparison of their strategies and logic.

\begin{algorithm}[t]
    \caption{AR2L Algorithm}
    \label{alg:ar2l}
    \textbf{Initialize}: packing policy $\pi_{\mathrm{pack}}$, permutation-based attacker $\pi_{\mathrm{perm}}$, mixture-dynamics model $\pi_{\mathrm{mix}}$, and their corresponding value functions $V^{\pi_{\mathrm{pack}}}$, $V^{\pi_{\mathrm{perm}}}$, $V^{\pi_{\mathrm{mix}}}$; \\
    \textbf{Input}: stationary item distribution $p_{b}$;  \\
    \textbf{Parameter}: robustness weight $\alpha$, number of observable items $N_B$;   \\
    \textbf{Output}: packing policy $\pi_{\mathrm{pack}}$;   \\
    \begin{algorithmic}[1] 
        \FOR{$i=0$ to $max\_iteration$}
        \STATE \texttt{\# train the permutation-based attacker}
        \STATE Reset the environment and observe $\bm{\mathrm{C}}_{0}, \bm{\mathrm{B}}_{0} \sim p_b$;
        \FOR{$t=0$ to $max\_step$}
        \STATE Get permuted item sequence: $\bm{\mathrm{B}}_{t}^{\prime} = \pi_{\mathrm{perm}}(\bm{\mathrm{C}}_{t}, \bm{\mathrm{B}}_{t})$;
        \STATE Get location for $b_{t,1}^{\prime}$: $l_{t} = \pi_{\mathrm{pack}}(\bm{\mathrm{C}}_{t}, \bm{\mathrm{B}}_{t}^{\prime})$;
        \STATE Pack $b_{t,1}^{\prime}$ into the bin, and observe $\bm{\mathrm{C}}_{t+1}, \bm{\mathrm{B}}_{t+1} \sim p_b$, reward $ r^{\mathrm{pack}}_{t}$, termination $d_{t}$;
        \IF{$d_{t} == \mathrm{True}$}
        \STATE Update $\pi_{\mathrm{perm}}$ on episode samples to maximize $-\sum r^{\mathrm{pack}}_{t}$;
        \ENDIF
        \ENDFOR

        \STATE \texttt{\# train the mixture-dynamics model and the packing policy}
        \STATE Reset the environment and observe $\bm{\mathrm{C}}_{0}, \bm{\mathrm{B}}_{0} \sim p_b$;
        \FOR{$t=0$ to $max\_step$}
        \STATE Get permuted item sequence: $\bm{\mathrm{B}}_{t}^{\prime\prime} = \pi_{\mathrm{mix}}(\bm{\mathrm{C}}_{t}, \bm{\mathrm{B}}_{t})$;
        \STATE Get location for $b_{t,1}^{\prime\prime}$: $l_{t} = \pi_{\mathrm{pack}}(\bm{\mathrm{C}}_{t}, \bm{\mathrm{B}}_{t}^{\prime\prime})$;
        \STATE Pack $b_{t,1}^{\prime\prime}$ into the bin, and observe $\bm{\mathrm{C}}_{t+1}, \bm{\mathrm{B}}_{t+1} \sim p_b$, reward $ r^{\mathrm{pack}}_{t}$, termination $d_{t}$;
        \IF{$d_{t} == \mathrm{True}$}
        \STATE Update $\pi_{\mathrm{mix}}, V^{\pi_{\mathrm{mix}}}$ on episode samples to maximize \\
        $\sum r^{\mathrm{pack}}_{t} - (D_{KL}(\pi_{\mathrm{mix}}||\mathbb{1}_{\{x=b_{t+1,1}})+\alpha D_{KL}(\pi_{\mathrm{mix}}||\pi_{\mathrm{perm}}))$;
        \STATE Update $\pi_{\mathrm{pack}}, V^{\pi_{\mathrm{pack}}}$ on episode samples to maximize $\sum r^{\mathrm{pack}}_{t}$;
        \ENDIF
        \ENDFOR
        \ENDFOR
    \end{algorithmic}
\end{algorithm}

We reproduced several learning-based algorithms following their original designs: AR2L, TAP-Net++, and PackE are based on the pseudocode in Algorithm~\ref{alg:ar2l}, Algorithm~\ref{alg:tapnet}, and Algorithm~\ref{alg:packe}, respectively. 

PCT and CDRL were implemented from the source codes at 
\href{https://github.com/alexfrom0815/Online-3D-BPP-PCT}{Online-3D-BPP-PCT} and 
\href{https://github.com/alexfrom0815/Online-3D-BPP-DRL}{Online-3D-BPP-DRL}; pseudocode is omitted.

The pseudocode for eight heuristic algorithms (HM, LASH, MACS, Random, OnlineBPH, DBL, BR, SDF Packing) is shown in Algorithm~\ref{alg:heightmapmin}-Algorithm~\ref{alg:sdfpacking}.

\begin{algorithm}
\caption{TAP-Net++ Algorithm}
\label{alg:tapnet}
\begin{algorithmic}[1]
\STATE \textbf{Input:} 
\STATE $\quad \mathcal{B} = \{ \mathbf{B}_j \}_{j=1}^N$ : Object states ($\mathbf{B}_j = [\mathbf{b}_j, \mathbf{p}_j]$)
\STATE $\quad \mathcal{E} = \{ \mathbf{E}_k \}_{k=1}^M$ : EMS candidates ($\mathbf{E}_k = [c_x, c_y, c_z, d_x, d_y, d_z]$)
\STATE \textbf{Output:} Optimal pair $(j^*, k^*)$

\STATE \textbf{Source Encoding:}
\FOR{$j = 1$ to $N$}
    \STATE $\mathbf{s}_j \leftarrow \text{ObjectEncoder}(\mathbf{B}_j)$
    \STATE $\mathbf{s}_j \leftarrow \mathbf{s}_j + \text{PositionalEncoding}(j)$
\ENDFOR
\STATE $\mathcal{S} \leftarrow \text{Attention}(\{\mathbf{s}_j\}_{j=1}^N)$  \COMMENT{Object features $N \times D$}

\STATE \textbf{Target Encoding:}
\FOR{$k = 1$ to $M$}
    \STATE $\mathbf{t}_k \leftarrow \text{MLP}_{\text{ems}}(\mathbf{E}_k)$
\ENDFOR
\STATE $\mathcal{T} \leftarrow \text{Attention}(\{\mathbf{t}_k\}_{k=1}^M)$  \COMMENT{EMS features $M \times D$}

\STATE \textbf{Match Scoring:}
\FOR{$j = 1$ to $N$}
\FOR{$k = 1$ to $M$}
    \STATE $\mathbf{M}[j,k] \leftarrow \mathbf{s}_j^\top \mathbf{t}_k$  \COMMENT{Matching score}
    \STATE $\mathbf{F}[j,k] \leftarrow \sigma(\text{MLP}_f(\mathbf{s}_j)^\top \text{MLP}_f(\mathbf{t}_k))$  \COMMENT{Feasibility mask}
    \STATE $\mathbf{V}[j,k] \leftarrow \mathbb{I}[\mathbf{b}_j \leq \mathbf{d}_k \text{ and accessible}]$  \COMMENT{Validity mask}
\ENDFOR
\ENDFOR

\STATE $\mathbf{P} \leftarrow \text{softmax}(\mathbf{M} \odot \mathbf{F} \odot \mathbf{V})$  \COMMENT{Pair probabilities}
\STATE $(j^*, k^*) \leftarrow \arg\max_{(j,k)} \mathbf{P}[j,k]$

\RETURN $(j^*, k^*)$
\end{algorithmic}
\end{algorithm}

\begin{algorithm}
\caption{PackE Algorithm}
\label{alg:packe}
\begin{algorithmic}[1]
\STATE \textbf{Input:} 
\STATE $\quad$ Bin dimensions $(L, W, H)$ \\
$\quad$ Object dimensions $(l, w, h)$ \\
$\quad$ Current bin heightmap $\mathcal{H}$

\STATE \textbf{Output:} Heuristic positions $\mathbf{E} \in \{0,1\}^{L \times W}$

\STATE Initialize $\mathbf{E} \gets \mathbf{0}$ \COMMENT{All zeros matrix}

\STATE \textbf{1. First Fit:}
\STATE \quad Find first empty $(i,j)$ in row-major order
\STATE \quad $\mathbf{E}[i,j] \gets 1$

\STATE \textbf{2. Floor Building:}
\STATE \quad Find $(i^*,j^*) = \text{lowest position in } \mathcal{H}$
\STATE \quad $\mathbf{E}[i^*,j^*] \gets 1$

\STATE \textbf{3. Column Building:}
\STATE \quad Find $(i^*,j^*) = \text{highest position in } \mathcal{H}$
\STATE \quad $\mathbf{E}[i^*,j^*] \gets 1$

\STATE \textbf{4. Extreme Points:}
\FOR{each placed object}
\STATE \quad Add right corner $(x+l, y)$
\STATE \quad Add front corner $(x, y+w)$
\STATE \quad $\mathbf{E}[\text{corners}] \gets 1$
\ENDFOR

\STATE \textbf{Filter invalid positions:}
\FOR{each $(i,j)$ where $\mathbf{E}[i,j] = 1$}
\IF{object doesn't fit at $(i,j)$}
\STATE $\mathbf{E}[i,j] \gets 0$
\ENDIF
\ENDFOR

\RETURN $\mathbf{E}$
\end{algorithmic}
\end{algorithm}

\begin{algorithm}[H]
\caption{HM}
\label{alg:heightmapmin}
\begin{algorithmic}[1]
\FOR{each position $(i,j)$ in bin}
    \STATE $score \gets i + j + 100 \times \sum \mathcal{S}[i:i+l, j:j+w]$
    \STATE Keep position with minimal score
\ENDFOR
\STATE \textbf{Output:} best position
\end{algorithmic}
\end{algorithm}

\begin{algorithm}[H]
\caption{LASH}
\label{alg:lash}
\begin{algorithmic}[1]
\FOR{each candidate space in $\mathcal{S}$}
    \STATE Compute new surface area after placement
    \STATE Keep placement with minimal surface increase
\ENDFOR
\STATE \textbf{Output:} best position
\end{algorithmic}
\end{algorithm}

\begin{algorithm}[H]
\caption{MACS}
\label{alg:macs}
\begin{algorithmic}[1]
\FOR{each candidate space in $\mathcal{S}$}
    \STATE Compute remaining space after placement
    \STATE Keep placement with maximal remaining space
\ENDFOR
\STATE \textbf{Output:} best position
\end{algorithmic}
\end{algorithm}

\begin{algorithm}[H]
\caption{OnlineBPH}
\label{alg:onlinebph}
\begin{algorithmic}[1]
\STATE Sort spaces by depth (lowest first)
\STATE Select first feasible position
\STATE \textbf{Output:} selected position
\end{algorithmic}
\end{algorithm}

\begin{algorithm}[H]
\caption{DBL}
\label{alg:dbl}
\begin{algorithmic}[1]
\FOR{each position $(i,j)$ in bin}
    \STATE $z \gets$ height at $(i,j)$
    \STATE $score \gets i + j + 100 \times z$
    \STATE Keep position with minimal score
\ENDFOR
\STATE \textbf{Output:} best position
\end{algorithmic}
\end{algorithm}

\begin{algorithm}[H]
\caption{BR}
\label{alg:br}
\begin{algorithmic}[1]
\FOR{each candidate space in $\mathcal{S}$}
    \STATE Compute space utilization score
    \STATE Count items that could fit in the space
    \STATE Keep space with the highest utility score
\ENDFOR
\STATE \textbf{Output:} best position
\end{algorithmic}
\end{algorithm}

\begin{algorithm}[H]
\caption{SDF Packing}
\label{alg:sdfpacking}
\begin{algorithmic}[1]
\FOR{each position $(i,j)$ and rotation}
    \STATE Compute TSDF + rotation penalty + height
    \STATE Keep position with minimal total
\ENDFOR
\STATE \textbf{Output:} best position
\end{algorithmic}
\end{algorithm}

\subsection{MORE VISUALIZED RESULTS}
\begin{figure*}[htbp]
    \centering
    \includegraphics[width=0.83\textwidth,keepaspectratio]{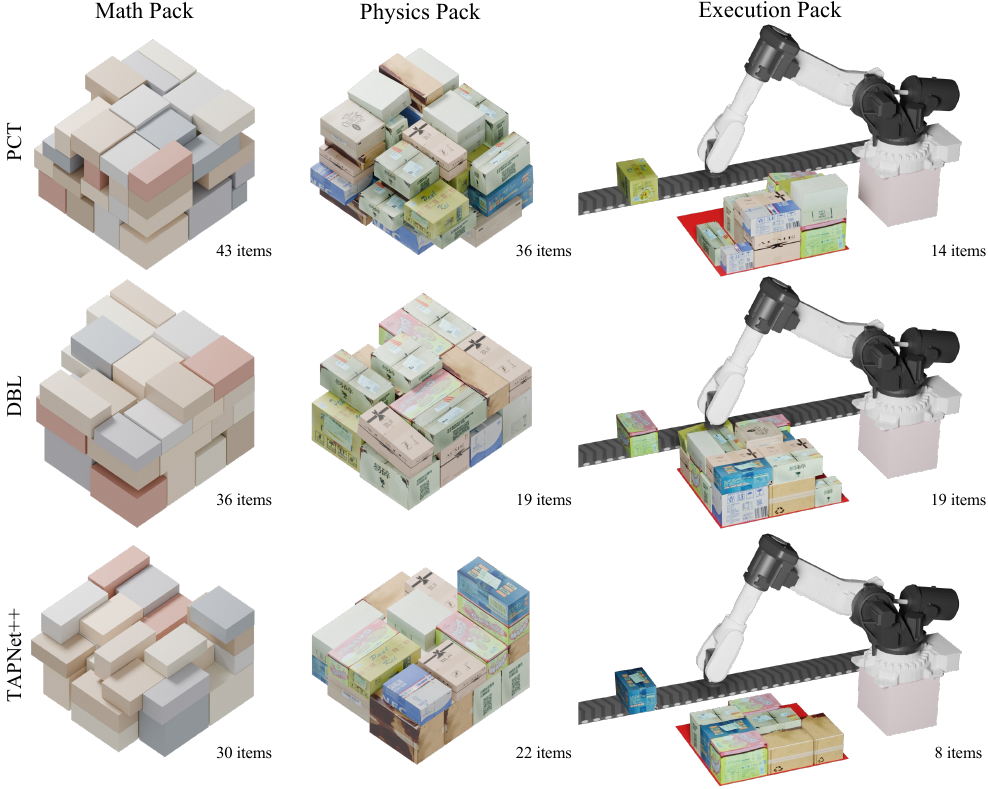}
    \caption{Visual placement results from one test of the Supplier Dataset.}
    \label{fig:pg}
\end{figure*}

\begin{figure*}[htbp]
    \centering
    \includegraphics[width=0.83\textwidth,keepaspectratio]{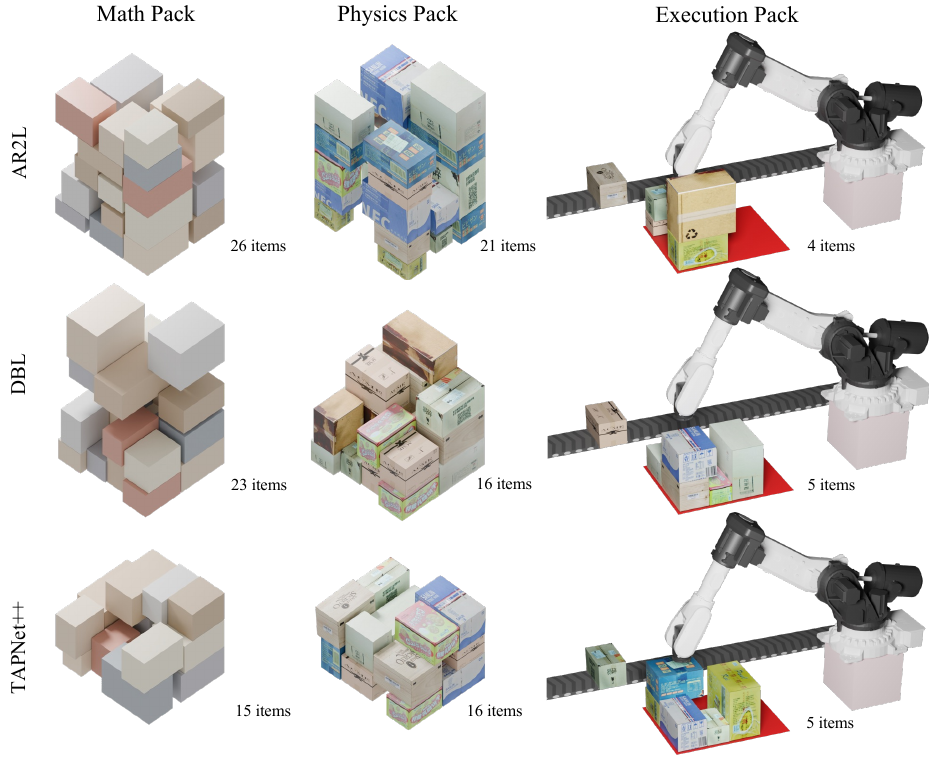}
    \caption{Visual placement results from one test of the Consumer Dataset.}
    \label{fig:deli}
\end{figure*}

\begin{figure*}[htbp]
    \centering
    \includegraphics[width=0.8\textwidth,keepaspectratio]{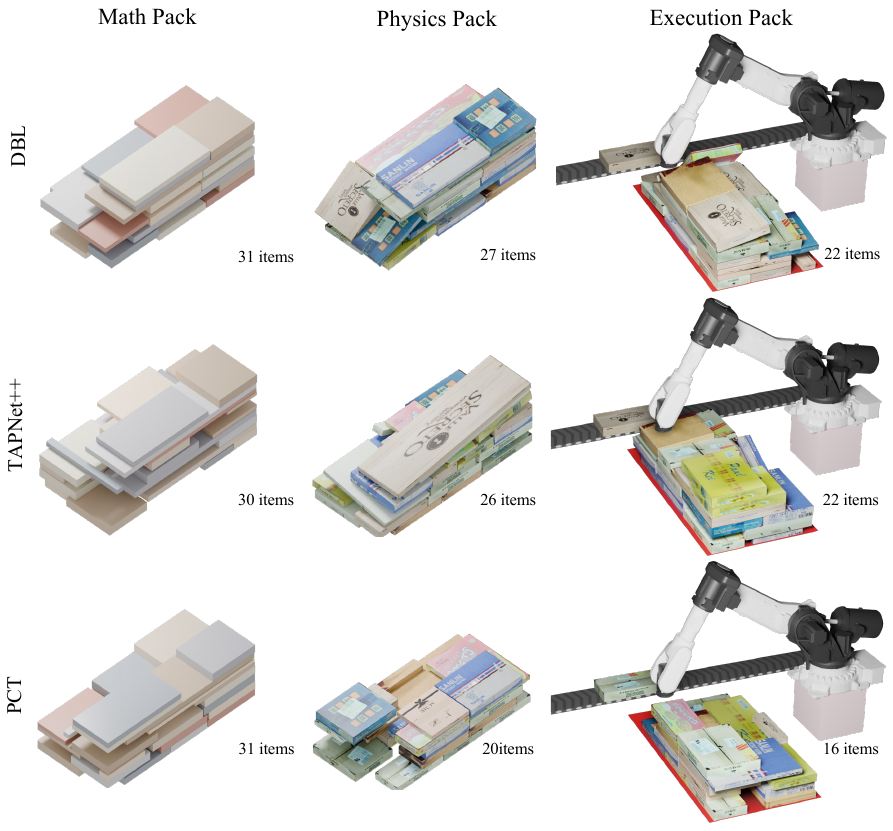}
    \caption{Visual placement results from one test of the Long Board Dataset.}
    \label{fig:opai}
\end{figure*}

We provide additional visualization results on all three industrial datasets under the three experimental settings. Specifically, we show representative packing results of several leading algorithms across Setting~1, Setting~2, and Setting~3. The galleries are presented in Figure~\ref{fig:pg}–Figure~\ref{fig:opai}, covering the Supplier Dataset, Consumer Dataset, and Long Board Dataset under different settings. In total, figures are included to provide an intuitive view of how algorithms perform under varying geometric and physical constraints.

\subsection{MORE DETAIL RESULTS}
This section presents the full evaluation results of all algorithms on the three datasets (\textit{Repetitive}, \textit{Diverse}, and \textit{Long Board}) under three test settings (\textit{Math Pack}, \textit{Physics Pack}, and \textit{Execution Pack}). Tables~\ref{tab:pg setting12}--\ref{tab:opai setting3} report detailed metrics, providing a complete view of algorithm performance across datasets and scenarios. Metric abbreviations: 
Uti. = Space Utilization, 
Static Stab. = Static Stability, 
Trajectory Len. = Trajectory Length,
Collapsed Plac. = Collapsed Placement, 
Dangerous Oper. = Dangerous Operation, 
Occ. = Occupancy, 
Deci. Time = Decision Time, 
Local Stab. = Local Stability.

\begin{table*}[t]
\centering
\caption{Performance of the Repetitive Dataset on Math Pack and Physics Pack.}
\scalebox{0.85}{
\setlength{\tabcolsep}{6 pt}
\renewcommand{\arraystretch}{1.0}

\begin{tabular}{@{}c|c@{}}
\hline
\multicolumn{1}{c|}{\textbf{Math Pack}} & \multicolumn{1}{c}{\textbf{Physics Pack}} \\
\hline
\begin{tabular}{c|ccc}
\hline
Method & \makecell{Uti. $\uparrow$} & \makecell{Occ. $\uparrow$} & \makecell{Deci. \\ Time $\downarrow$} \\
\hline
AR2L & 0.673 & 0.898 & 1.499\\
PCT & 0.659 & 0.895 & 0.546 \\
LSAH & 0.619 & 0.908 & 0.004  \\
DBL & 0.617 & 0.911 & 0.402  \\
BR & 0.587 & 0.873 & 0.005  \\
SDFPack & 0.586 & 0.834 & 4.249  \\
TAP-Net++ & 0.579 & 0.915 & 0.278  \\
HM & 0.579 & 0.960 & 0.763  \\
OnlineBPH & 0.5 & 0.739 & $\bm{0.003}$  \\
MACS & 0.445 & 0.648 & 6.759  \\
CDRL & 0.425 & 0.826 & 0.268  \\
PackE & 0.312 & $\bm{0.965}$ & 0.113  \\
\hline
\end{tabular}
& 
\begin{tabular}{c|ccccc}
\hline
Method & \makecell{Uti. $\uparrow$} & \makecell{Occ. $\uparrow$} & \makecell{Deci. \\ Time $\downarrow$} & \makecell{Local \\ Stab. $\downarrow$} & \makecell{Static \\ Stab. $\uparrow$} \\
\hline
HM & 0.398 & $\bm{0.965}$ & 0.817 & 0.026 & 0.347 \\
PCT & 0.392 & 0.942 & 0.617 & 0.010 & 0.386 \\
DBL & 0.385 & 0.947 & 0.632 & 0.009 & 0.400 \\
TAP-Net++ & 0.379 & 0.948 & 0.809 & $\bm{0.006}$ & 0.415 \\
LSAH & 0.373 & 0.943 & 0.005 & 0.008 & 0.410 \\
PackE & 0.353 & $\bm{0.965}$ & 0.105 & 0.013 & 0.410 \\
AR2L & 0.229 & 0.958 & 1.523 & $\bm{0.006}$ & $\bm{0.428}$ \\
BR & 0.222 & 0.927 & 0.005 & 0.011 & 0.381 \\
CDRL & 0.21 & 0.888 & 2.716 & 0.013 & 0.426 \\
SDFPack & 0.15 & 0.918 & 7.537 & 0.009 & 0.404 \\
OnlineBPH & 0.116 & 0.941 & $\bm{0.001}$ & 0.007 & 0.407 \\
MACS & 0.07 & 0.911 & 1.244 & 0.019 & 0.401 \\
\hline
\end{tabular} \\
\hline
\end{tabular}
} 
\label{tab:pg setting12}
\end{table*}

\begin{table*}[t]
\centering
\caption{Performance of the Repetitive Dataset on Execution Pack.}
\scalebox{0.85}{
\begin{tabular}{c|ccccccccc}
\hline
Method & \makecell{Uti. $\uparrow$} 
       & \makecell{Occ. $\uparrow$} 
       & \makecell{Deci. \\ Time $\downarrow$} 
       & \makecell{Local \\ Stab. $\downarrow$} 
       & \makecell{Static \\ Stab. $\uparrow$}  
       & \makecell{Trajectory \\ Len. $\downarrow$} 
       & \makecell{Collapsed \\ Plac. $\downarrow$} 
       & \makecell{Dangerous \\ Oper. $\downarrow$} \\
\hline
PCT          & $\bm{0.383}$ & 0.944 & 0.131 & 0.012 & 0.383 & 2.38  & $\bm{0.033}$ & 0.028 \\
DBL          & 0.359 & 0.948 & 0.611 & 0.013 & 0.410 & 2.253 & 0.041 & $\bm{0.015}$ \\
TAP-Net++     & 0.335 & 0.952 & 0.793 & 0.012 & 0.420 & 2.371 & 0.068 & 0.065 \\
LSAH         & 0.304 & 0.945 & 0.004 & 0.033 & 0.398 & 2.313 & 0.108 & 0.052 \\
AR2L         & 0.226 & 0.959 & 1.518 & 0.027 & 0.430 & 2.351 & 0.073 & 0.018 \\
PackE        & 0.169 & 0.965 & 0.166 & 0.228 & 0.428 & 2.200 & 0.302 & 0.098 \\
BR           & 0.121 & 0.942 & 0.004 & 0.101 & 0.371 & 2.454 & 0.248 & 0.096 \\
OnlineBPH    & 0.118 & 0.940 & $\bm{0.001}$ & $\bm{0.009}$ & 0.414 & 2.790 & 0.157 & 0.034 \\
SDFPack      & 0.098 & 0.929 & 12.672 & 0.151 & 0.370 & 2.306 & 0.385 & 0.234 \\
HM & 0.083 & $\bm{0.966}$ & 0.495 & 0.415 & 0.417 & 2.475 & 0.709 & 0.074 \\
CDRL         & 0.080 & 0.927 & 0.472 & 0.460 & $\bm{0.468}$ & $\bm{2.104}$ & 0.676 & 0.037 \\
MACS         & 0.047 & 0.915 & 1.244 & 0.302 & 0.423 & 2.744 & 0.653 & $\bm{0.000}$ \\
\hline
\end{tabular}
}
\label{tab:pg setting3}
\end{table*}

\begin{table*}[t]
\centering
\caption{Performance of the Diverse Dataset on Math Pack and Physics Pack.}
\scalebox{0.85}{
\setlength{\tabcolsep}{6pt}
\renewcommand{\arraystretch}{1.0}
\begin{tabular}{@{}c|c@{}}
\hline
\multicolumn{1}{c|}{\textbf{Math Pack}} & \multicolumn{1}{c}{\textbf{Physics Pack}} \\
\hline
\begin{tabular}{c|ccc}
\hline
Method & \makecell{Uti. $\uparrow$} & \makecell{Occ. $\uparrow$} & \makecell{Deci. \\ Time $\downarrow$} \\
\hline
PCT          & $\bm{0.668}$   & 0.897           & 0.557                 \\
AR2L         & 0.655          & 0.878           & 1.499                 \\
LSAH         & 0.569          & 0.885           & 0.002                 \\
DBL          & 0.549          & 0.886           & 0.094                 \\
SDFPack      & 0.543          & 0.849           & 5.754                 \\
BR           & 0.524          & 0.836           & 0.002          \\
CDRL         & 0.501          & 0.834           & 0.132                 \\
OnlineBPH    & 0.474          & 0.768           & $\bm{0.001}$         \\
HM & 0.423          & $\bm{0.943}$    & 0.391                 \\
MACS         & 0.352          & 0.687           & 5.623                 \\
PackE        & 0.323          & 0.898           & 0.109                 \\
TAP-Net++     & 0.263          & 0.936           & 0.233                 \\
\hline
\end{tabular}
& 
\begin{tabular}{c|ccccc}
\hline
Method & \makecell{Uti. $\uparrow$} & \makecell{Occ. $\uparrow$} & \makecell{Deci. \\ Time $\downarrow$} & \makecell{Local \\ Stab. $\downarrow$} & \makecell{Static \\ Stab. $\uparrow$} \\
\hline
PCT          & $\bm{0.502}$ & 0.921 & 0.585 & 0.009 & 0.436\\
AR2L         & 0.483 & 0.867 & 1.510 & 0.014 & 0.446  \\
CDRL         & 0.375 & 0.871 & 2.654 & 0.010 & 0.435  \\
HM & 0.330 & 0.965 & 0.426 & 0.010 & 0.426  \\
LSAH         & 0.285 & 0.947 & 0.003 & 0.009 & 0.424  \\
TAP-Net++     & 0.270 & 0.962 & 0.355 & 0.004 & 0.426  \\
SDFPack      & 0.259 & 0.909 & 6.762 & 0.014 & 0.445  \\
BR           & 0.258 & 0.895 & 0.003 & 0.024 & 0.428  \\
DBL          & 0.257 & 0.956 & 0.279 & 0.006 & 0.427  \\
PackE        & 0.171 & $\bm{0.986}$ & 0.240 & $\bm{0.001}$ & $\bm{0.478}$  \\
MACS         & 0.134 & 0.880 & 2.760 & 0.029 & 0.449  \\
OnlineBPH    & 0.120 & 0.975 & $\bm{0.001}$ & 0.002 & 0.458  \\
\hline
\end{tabular} \\
\hline
\end{tabular}
}
\label{tab:deli setting12}
\end{table*}

\begin{table*}[t]
\centering
\caption{Performance of the Diverse Dataset on Execution Pack.}
\scalebox{0.85}{
\begin{tabular}{c|ccccccccc}
\hline
Method & \makecell{Uti. $\uparrow$} 
       & \makecell{Occ. $\uparrow$} 
       & \makecell{Deci. \\ Time $\downarrow$} 
       & \makecell{Local \\ Stab. $\downarrow$} 
       & \makecell{Static \\ Stab. $\uparrow$}  
       & \makecell{Trajectory \\ Len. $\downarrow$} 
       & \makecell{Collapsed \\ Plac. $\downarrow$} 
       & \makecell{Dangerous \\ Oper. $\downarrow$}  \\
\hline
AR2L         & $\bm{0.240}$ & 0.918 & 1.487 & 0.294 & 0.464 & 2.109 & 0.538 & 0.066  \\
CDRL         & 0.208 & 0.931 & 0.267 & 0.082 & 0.460 & 2.347 & $\bm{0.256}$ & 0.055  \\
DBL          & 0.183 & 0.968 & 0.503 & 0.051 & 0.451 & 2.206 & 0.478 & 0.046  \\
HM & 0.143 & 0.980 & 0.317 & 0.252 & 0.460 & 2.221 & 0.651 & 0.031  \\
TAP-Net++     & 0.141 & 0.976 & 0.553 & 0.073 & 0.444 & 2.115 & 0.544 & 0.132  \\
PCT          & 0.126 & 0.964 & 0.042 & 0.192 & 0.472 & $\bm{1.989}$ & 0.452 & 0.010  \\
MACS         & 0.125 & 0.885 & 2.577 & 0.455 & 0.451 & 2.378 & 0.499 & $\bm{0.006}$  \\
SDFPack      & 0.100 & 0.940 & 8.067 & 0.362 & 0.416 & 2.148 & 0.766 & 0.191  \\
OnlineBPH    & 0.097 & $\bm{0.981}$ & $\bm{0.002}$ & 0.029 & 0.476 & 2.232 & 0.552 & 0.033  \\
LSAH         & 0.091 & 0.967 & $\bm{0.002}$ & 0.154 & 0.464 & 2.231 & 0.652 & 0.036  \\
PackE        & 0.078 & 0.918 & 0.498 & $\bm{0.002}$ & $\bm{0.478}$ & 2.495 & 0.408 & 0.007  \\
BR           & 0.067 & 0.942 & 0.004 & 0.396 & 0.458 & 2.212 & 0.815 & 0.057 \\
\hline
\end{tabular}
}
\label{tab:deli setting3}
\end{table*}

\begin{table*}[t]
\centering
\caption{Performance of the Long Board Dataset on Math Pack and Physics Pack.}
\scalebox{0.85}{
\setlength{\tabcolsep}{6pt}
\renewcommand{\arraystretch}{1.0}
\begin{tabular}{@{}c|c@{}}
\hline
\multicolumn{1}{c|}{\textbf{Math Pack}} & \multicolumn{1}{c}{\textbf{Physics Pack}} \\
\hline
\begin{tabular}{c|ccc}
\hline
Method & \makecell{Uti. $\uparrow$} & \makecell{Occ. $\uparrow$} & \makecell{Deci. \\ Time $\downarrow$} \\
\hline
PCT          & $\bm{0.571}$ & 0.771           & 0.517                 \\
AR2L         & 0.538        & 0.716           & 1.180                \\
DBL          & 0.520        & 0.769           & 0.400                 \\
TAP-Net++     & 0.519        & 0.724           & 0.033          \\
LSAH         & 0.474        & 0.686           & $\bm{0.002}$          \\
CDRL         & 0.451        & 0.699           & 0.305                 \\
OnlineBPH    & 0.428        & 0.676           & $\bm{0.002}$          \\
BR           & 0.363        & 0.627           & 0.003                 \\
HM & 0.355        & $\bm{0.787}$    & 0.799                 \\
SDFPack      & 0.274        & 0.619           & 13.343               \\
PackE        & 0.240        & 0.647           & 0.140                 \\
MACS         & 0.232        & 0.610           & 12.461               \\
\hline
\end{tabular}
& 
\begin{tabular}{c|ccccc}
\hline
Method & \makecell{Uti. $\uparrow$} & \makecell{Occ. $\uparrow$} & \makecell{Deci. \\ Time $\downarrow$} & \makecell{Local \\ Stab. $\downarrow$} & \makecell{Static \\ Stab. $\uparrow$} \\
\hline
DBL          & $\bm{0.349}$ & 0.815 & 0.744 & 0.041 & 0.259 \\
TAP-Net++     & 0.314 & 0.801 & 0.091 & 0.043 & 0.259  \\
PCT          & 0.290 & 0.802 & 0.527 & $\bm{0.034}$ & $\bm{0.371}$  \\
CDRL         & 0.286 & 0.759 & 0.064 & 0.046 & 0.268  \\	
OnlineBPH    & 0.277 & 0.745 & $\bm{0.002}$ & 0.036 & 0.222 \\
AR2L         & 0.265 & 0.792 & 1.183 & 0.043 & 0.334  \\
HM & 0.255 & $\bm{0.831}$ & 0.980 & 0.046 & 0.306  \\
LSAH         & 0.254 & 0.748 & $\bm{0.002}$ & 0.045 & 0.244  \\
BR           & 0.169 & 0.729 & 0.003 & 0.057 & 0.263  \\
PackE        & 0.162 & 0.687 & 0.082 & 0.043 & 0.208 \\
SDFPack      & 0.155 & 0.732 & 14.225 & 0.058 & 0.289 \\
MACS         & 0.135 & 0.738 & 4.636 & 0.049 & 0.266  \\
\hline
\end{tabular} \\
\hline
\end{tabular}
}
\label{tab:opai setting12}
\end{table*}

\begin{table*}[t]
\centering
\caption{Performance of the Long Board Dataset on Execution Pack.}
\scalebox{0.85}{
\wzf{
\begin{tabular}{c|ccccccccc}
\hline
Method & \makecell{Uti. $\uparrow$} 
       & \makecell{Occ. $\uparrow$} 
       & \makecell{Deci. \\ Time $\downarrow$} 
       & \makecell{Local \\ Stab. $\downarrow$} 
       & \makecell{Static \\ Stab. $\uparrow$}  
       & \makecell{Trajectory \\ Len. $\downarrow$} 
       & \makecell{Collapsed \\ Plac. $\downarrow$} 
       & \makecell{Dangerous \\ Oper. $\downarrow$}  \\
\hline
DBL          & $\bm{0.285}$ & 0.837 & 0.722 & 0.045 & 0.282 & 3.195 & $\bm{0.118}$ & 0.024 \\
CDRL         & 0.252 & 0.783 & 0.261 & 0.048 & 0.285 & 3.077 & 0.138 & 0.027 \\
PCT          & 0.244 & 0.825 & 0.029 & $\bm{0.042}$ & $\bm{0.384}$ & 3.196 & 0.147 & 0.013 \\
OnlineBPH    & 0.240 & 0.767 & $\bm{0.002}$ & $\bm{0.042}$ & 0.242 & 3.019 & 0.143 & 0.013 \\
TAP-Net++    & 0.225 & 0.852 & 0.285 & 0.053 & 0.324 & 3.277 & 0.131 & 0.049 \\
AR2L         & 0.221 & 0.812 & 0.798 & 0.055 & 0.346 & 2.753 & 0.214 & 0.021 \\
HM           & 0.214 & $\bm{0.856}$ & 1.007 & 0.054 & 0.333 & 3.270 & 0.168 & 0.028 \\
LSAH         & 0.187 & 0.774 & $\bm{0.002}$ & 0.059 & 0.273 & 3.230 & 0.218 & 0.041 \\
PackE        & 0.146 & 0.706 & 0.152 & 0.064 & 0.217 & $\bm{2.010}$ & 0.210 & 0.049 \\
BR           & 0.124 & 0.769 & $\bm{0.002}$ & 0.071 & 0.292 & 3.752 & 0.314 & 0.007 \\
MACS         & 0.123 & 0.748 & 4.304 & 0.068 & 0.256 & 3.782 & 0.276 & $\bm{0.004}$ \\
SDFPack      & 0.119 & 0.768 & 14.295 & 0.070 & 0.289 & 3.787 & 0.327 & 0.009 \\
\hline
\end{tabular}
}
}
\label{tab:opai setting3}
\end{table*}

\end{document}